\documentclass[pdflatex,sn-basic,oneside]{sn-jnl}

\usepackage{wrapfig}
\usepackage{graphicx}
\usepackage{multirow}
\usepackage{amsmath,amssymb,amsfonts}
\usepackage{mathrsfs}
\usepackage[title]{appendix}
\usepackage{xcolor}
\usepackage{mathtools}
\usepackage{textcomp}
\usepackage{manyfoot}
\usepackage{booktabs}
\usepackage{algorithm}
\usepackage{algorithmicx}
\usepackage{algpseudocode}
\usepackage{listings}
\usepackage{lipsum}
\usepackage{tikz}
\usepackage{tabto}
\usepackage{tabularx}
\usepackage[figuresright]{rotating}
\usepackage{array}
\usepackage[capitalise,noabbrev,nameinlink]{cleveref}

\makeatletter

\newcommand{\Rmnum}[1]{\expandafter\@slowromancap\romannumeral #1@}
\newcommand{\RaggedRight}{\rightskip\@flushglue\let\\\@centercr}
\renewcommand{\orcid}[1]{%
  \raisebox{-0.1ex}{%
    \includegraphics[height=2ex]{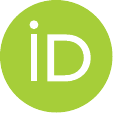}%
  }%
}
\makeatother

\begin{document}

\title{SAMP-HDRL: Segmented Allocation with Momentum-Adjusted Utility for Multi-agent Portfolio Management via Hierarchical Deep Reinforcement Learning}

\author[1]{Xiaotian Ren\orcid{0000-0001-7706-6268}}
\author[2]{Nuerxiati Abudurexiti\orcid{0009-0002-6052-0051}}
\author[1]{Zhengyong Jiang\orcid{0000-0001-8873-4073}}
\author[1]{Angelos Stefanidis\orcid{0000-0002-4703-8765}}

\author*[1]{Hongbin Liu\orcid{0000-0002-7398-9193}}
\email{Hongbin.Liu@xjtlu.edu.cn}

\author*[1]{Jionglong Su\orcid{0000-0001-5360-6493}}
\email{Jionglong.Su@xjtlu.edu.cn}

\affil[1]{School of AI and Advanced Computing,
XJTLU Entrepreneur College (Taicang),
Xi’an Jiaotong-Liverpool University,
Suzhou 215123, China}

\affil[2]{Department of Management Science and Engineering,
School of Economics and Management,
Xinjiang University,
Urumqi 830000, China}

\abstract{
Portfolio optimization in non-stationary markets is challenging due to regime shifts, dynamic correlations, and the limited interpretability of deep reinforcement learning (DRL) policies. We propose a Segmented Allocation with Momentum-Adjusted Utility for Multi-agent Portfolio Management via Hierarchical Deep Reinforcement Learning (SAMP-HDRL). The framework first applies dynamic asset grouping to partition the market into high-quality and ordinary subsets. An upper-level agent extracts global market signals, while lower-level agents perform intra-group allocation under mask constraints. A utility-based capital allocation mechanism integrates risky and risk-free assets, ensuring coherent coordination between global and local decisions.  backtests across three market regimes (2019--2021) demonstrate that SAMP-HDRL consistently outperforms nine traditional baselines and nine DRL benchmarks under volatile and oscillating conditions. Compared with the strongest baseline, our method achieves at least 5\% higher Return, 5\% higher Sharpe ratio, 5\% higher Sortino ratio, and 2\% higher Omega ratio, with substantially larger gains observed in turbulent markets. Ablation studies confirm that upper--lower coordination, dynamic clustering, and capital allocation are indispensable to robustness. SHAP-based interpretability further reveals a complementary ``diversified + concentrated'' mechanism across agents, providing transparent insights into decision-making.  Overall, SAMP-HDRL embeds structural market constraints directly into the DRL pipeline, offering improved adaptability, robustness, and interpretability in complex financial environments.
}

\section*{Highlights}
\begin{itemize}
    \item SAMP-HDRL integrates dynamic asset grouping, hierarchical agent coordination, and utility-based capital allocation to enhance portfolio robustness under non-stationary market conditions.

    \item Extensive backtests across three market regimes show that SAMP-HDRL outperforms nine traditional and nine DRL baselines, delivering at least 5\% improvements in Return and major risk-adjusted metrics.

    \item SHAP-based analysis uncovers a complementary “diversified + concentrated’’ decision pattern across agent layers, providing transparent and interpretable insights into hierarchical DRL portfolio behavior.
\end{itemize}


\maketitle

\section{Introduction}\label{Intro}

Portfolio management is a core component of investment strategy, aiming to maximize risk-adjusted returns while controlling exposure \cite{bib1}. It constitutes a foundational problem at the intersection of finance, mathematical optimization, and computational intelligence. As financial markets become increasingly volatile and data complexity increases, traditional portfolio optimization strategies encounter significant limitations in terms of modeling capacity and adaptability \cite{bib2}. In recent years, Deep Reinforcement Learning (DRL) \cite{bib3} has emerged as a powerful paradigm for portfolio management due to its ability to model sequential decision-making under uncertainty \cite{bib57} and to capture complex non-linear dependencies among assets \cite{bib107}. As an advanced framework within machine learning, DRL integrates reinforcement learning principles with deep neural architectures, thereby enabling the design of strategies that optimize long-term, risk-adjusted returns in highly volatile financial markets. DRL enables an agent to continuously interact with its environment, iteratively updating its decision-making policy to maximize long-term returns and to adaptively realize optimal investment behaviors under varying market conditions \cite{bib5}. Benefiting from this reward-driven, self-adaptive optimization process, DRL exhibits substantial advantages in handling high-dimensional, dynamic, and nonlinear financial environments \cite{bib6}. Compared to traditional strategies, DRL-based portfolio management not only facilitates adaptive modeling of complex market dynamics but also offers enhanced capabilities to capture nonlinear relationships, temporal dependencies, and long-term return objectives \cite{bib5,bib6}, thereby establishing itself as a prominent research paradigm in intelligent investment systems.

In the domain of portfolio optimization, achieving effective allocation under dynamically evolving market conditions remains a fundamental challenge in intelligent systems research. Early studies primarily focus on holistic modeling of a single asset universe, emphasizing the use of more powerful and generalized neural architectures or advanced mathematical formulations to optimize portfolios \cite{bib7,bib8,bib9,bib10,bib11,bib12,bib13,bib14,bib15}. However, such approaches in high-dimensional spaces are highly prone to the curse of dimensionality, resulting in slow convergence, unstable training, or even divergence \cite{bib16}. To address this issue, researchers introduce clustering mechanisms to categorize assets and subsequently optimize portfolio weights through mathematical models, which subsequently improves performance \cite{bib17}. Nevertheless, these approaches generally suffer from a fundamental limitation: the clustering and optimization processes are decoupled. Since the clustering outcomes and the optimizer cannot be jointly trained, information cannot be effectively propagated, leading to constrained policy learning or suboptimal solutions \cite{bib18}. Moreover, most clustering procedures are static or rely on heuristic updates, which fail to capture the temporal dynamics of financial time series \cite{bib19}. Consequently, when regime shifts occur in the market, such models exhibit delayed or ineffective responses, resulting in severe drawdowns, uncontrolled risk exposure, and missed opportunities for potential profit \cite{bib20}. Leveraging this, some studies attempt to integrate clustering with deep reinforcement learning, applying DRL algorithms within clustered subsets to achieve preliminary coupling between clustering and DRL \cite{bib21,bib25}. However, clustering remains static and weakly integrated with the DRL pipeline, limiting the capacity to model dynamically changing market conditions \cite{bib22}. Furthermore, certain models introduce dynamic subset selection mechanisms, restricting investment to a subset of promising assets and reducing computational complexity by maintaining a representative subset of assets through online updates \cite{bib23,bib24}. Although online updating provides promising benefits, existing mechanisms predominantly rely on heuristic rules and do not incorporate differentiable end-to-end training frameworks, thereby limiting integration with deep learning or reinforcement learning and restricting scalability as well as generalization capability \cite{bib25}. To address these issues, researchers design a two-level reinforcement learning architecture in which the upper-level agent dynamically selects stocks while the lower-level agent optimizes transaction execution, thereby alleviating to some extent the challenges posed by high-dimensional action spaces \cite{bib26}. However, the stock selection process remains implicit \cite{bib26}, relying entirely on the policy network’s mapping from market states to actions and lacking explicit modeling of structural market shifts. Moreover, the inter-layer information flow does not constitute a genuine end-to-end feedback loop, which imposes inherent limitations on interpretability and adaptability to dynamic environments \cite{bib27}.

From a computational perspective, these approaches exhibit three major limitations. 

1. \textbf{Static or heuristic clustering lacks the ability to model the dynamic properties of time series.}
Many methods treat clustering as a one-off or low-frequency “static annotation,” \cite{bib21,bib25} or perform heuristic updates with fixed thresholds and simple rules, implicitly assuming that asset relationships and risk structures remain stationary within a given window \cite{bib23,bib24}. In reality, markets are highly dynamic, and such static clusters quickly become obsolete \cite{bib21}. The absence of dynamism leads to two consequences: (i) signal lag and attenuation—when regime shifts occur, outdated clusters or features lose discriminative power and strategies respond with delay \cite{bib28}; and (ii) risk mismatch—risk models constrained by outdated partitions underestimate true exposures, resulting in elevated risk \cite{bib29}. Moreover, this increases transaction costs, as misaligned rebalancing and frequent ineffective transactions cause excessive turnover. Methodologically, it is essential to introduce risk-sensitive dynamic clustering mechanisms and deeply couple such dynamic representations with portfolio optimization in order to achieve rapid and adaptive strategy updates in evolving environments.

2. \textbf{Clustering or subset selection is decoupled from the optimizer, preventing the formation of an end-to-end feedback loop.}
Approaches that rely on clustering or subset selection generally treat this procedure as a pre-processing stage, where the derived asset sets are subsequently passed to the optimizer or DRL module for portfolio weight learning \cite{bib17,bib23,bib24}. The two stages neither share parameters nor provide differentiable linkages; thus, the optimizer’s objective function or DRL reward signals cannot propagate backward to influence upstream clustering or filtering strategies. This disjunction creates both objective mismatch \cite{bib30} and distribution shift \cite{bib31}, leading to under-utilization of information \cite{bib31}, persistent errors \cite{bib30}, low sample efficiency \cite{bib32}, slow convergence \cite{bib33}, and high sensitivity to hyperparameters \cite{bib32}, with the risk of becoming trapped in local optima under complex non-convex objectives \cite{bib34}. Small perturbations upstream can also amplify downstream decisions, degrading stability and reproducibility \cite{bib35}. Without a closed loop, models cannot perform end-to-end uncertainty quantification and calibration, thereby weakening risk awareness and control \cite{bib36}. It is therefore necessary to design differentiable and feedback-aware coupling mechanisms that align asset clustering with portfolio objectives under a unified optimization paradigm.

3. \textbf{Subset updating suffers from limited scalability.} 
Rule-based dynamic subset selection cannot adaptively optimize with respect to rewards or losses \cite{bib23,bib24}, which easily introduces selection bias \cite{bib130}. The Hierarchical Reinforced Trader \cite{bib26}, designed with a two-level structure of “upper-level stock selection and lower-level execution,” illustrates how hierarchical designs can partially alleviate instability and exploration difficulties in high-dimensional action spaces. Such models are generally aligned with the paradigm of hierarchical reinforcement learning (HRL) \cite{bib48}, where complex decision-making tasks are decomposed into layered sub-problems to improve learning efficiency and robustness. However, the stock selection logic in these designs remains implicit, relying entirely on black-box policy networks mapping states to actions, and thus lacking explicit modeling of structural market changes \cite{bib38}. Moreover, commonly adopted alternating or staged training strategies hinder end-to-end information flow across layers, which not only exacerbates stability issues \cite{bib39} but also restricts scalability when applied to larger universes or higher-frequency trading environments. To address these challenges, it is essential to replace heuristic and black-box selection with explicit market-structure modeling, differentiable selection and allocation modules, and hierarchies that preserve informational consistency, while embedding capital allocation and risk control into a unified, utility-driven, learnable framework to enhance interpretability and scalability.

In summary, these three deficiencies collectively undermine generalization capability in real-world environments, delay responses to abrupt market transitions, and destabilize risk control. Consequently, dynamic modeling of non-stationary environments, systematic optimization around differentiable feedback loops, and scalable hierarchical architectures is not only critical for performance improvement but also a prerequisite for ensuring the practicality and reliability of knowledge-based systems in financial applications.

To address three major limitations in prevailing portfolio optimization techniques—static clustering, the lack of end-to-end integration between clustering and optimization, and reliance on heuristic subset mechanisms—we propose Segmented Allocation with momentum-adjusted utility for Multi-agent Portfolio management via Hierarchical Deep Reinforcement Learning (SAMP-HDRL), a knowledge-driven hierarchical DRL framework that integrates dynamic clustering with multi-agent decision making.In this framework, assets are dynamically partitioned into two groups through clustering. The upper-level agent operates on the complete asset universe, extracting global representations that capture inter-asset correlations and market-wide dynamics, and provides holistic guidance signals to inform subsequent decisions. The lower-level agents then focus on their respective dynamically assigned groups, allocating portfolio weights under masking constraints to ensure coherent intra-group optimization. An exponential utility function subsequently integrates the outputs of the lower-level agents with historical returns to determine segmented allocations across the two risky asset groups and the risk-free asset. On this basis, momentum adjustment and rebound detection are incorporated into the utility formulation to enhance robustness against persistent market trends and abrupt regime shifts.
By design, SAMP-HDRL not only adapts to dynamic changes in market structure but also achieves a principled integration of global feature modeling, localized optimization, and risk-sensitive capital allocation. In doing so, it provides a knowledge-driven and practically scalable solution to portfolio management, directly addressing the fragmentation and rigidity observed in conventional methodologies.

This study does not introduce new neural network architectures. This study concentrates on structural innovations within the reinforcement learning framework for portfolio management. The proposed design integrates dynamic asset grouping, coordinated multi-agent allocation, and risk-sensitive utility formulation. By emphasizing structural modeling over architectural modification, the framework ensures that performance improvements originate from principled design mechanisms, thereby enhancing interpretability, robustness, and scalability in practical financial applications.

This paper has four key contributions:

1. \textbf{Joint modeling of global market signals and localized asset-group decisions.}
Unlike conventional portfolio optimization methods that either rely on a single global model or isolate group-wise optimization, our framework first performs dynamic asset grouping, then applies an upper-level agent to extract global market representations, and finally deploys lower-level agents to allocate weights within their dynamically assigned clusters under mask constraints. This sequential design—grouping, global modeling, and local allocation—enhances representational richness and enables more precise allocation strategies, which is empirically reflected in improved adaptability and performance stability.

2. \textbf{Interpretable hierarchical decision mechanism built on dynamic classification.}
Leveraging dynamic asset classification, we design a hierarchical decision mechanism that eliminates the reliance on rule-based subset updating and implicit black-box selection. By explicitly modeling the grouping and allocation process within a unified learning framework, and by ensuring consistent information flow across layers, the proposed design mitigates instability caused by staged training. This approach enhances interpretability and provides stronger adaptability to structural market changes.

3. \textbf{An innovative momentum-adjusted utility function for segmented allocation.}
We advance the utility-based allocation principle by incorporating momentum adjustment \cite{bib103} and rebound detection \cite{bib108} into the capital allocation process, which jointly accounts for historical returns, regime dynamics, and risk-free assets. This enriched utility design enhances resilience against abrupt market transitions, offering a novel mechanism for risk-sensitive and knowledge-driven portfolio optimization.

4. \textbf{Consistent Outperformance in Diverse Markets}
SAMP-HDRL demonstrates consistent empirical performance under volatile and non-stationary market regimes. Compared with the strongest baselines, it achieves improvements of at least 5\%, 5\%, 5\%, and 2\% in Return, Sharpe Ratio, Sortino Ratio, and Omega Ratio, respectively. These results indicate that SAMP-HDRL not only enhances profitability but also delivers robust risk-adjusted performance, thereby underscoring both its theoretical contribution to reinforcement learning research and its practical significance for portfolio management.

The structure of this paper is as follows. Section 2 reviews the relevant literature and highlights recent developments in the domain of portfolio optimization and intelligent systems. Section 3 outlines the proposed methodology, covering the mathematical foundations, the design of the learning environment, and the architecture of the policy network. Section 4 presents the experimental backtests, where the performance of our framework is evaluated using portfolio value, Sharpe ratio, Sortino ratio, and Omega ratio against traditional strategies and learning-based approaches. Section 5 concludes the study and discusses possible directions for future research.

\section{Related Work of Portfolio Management}\label{Lit-rev}
In the domain of portfolio optimization, prior research primarily focuses on traditional online portfolio strategies. The Constant Rebalanced Portfolio (CRP)~\cite{bib71,bib72} and Universal Portfolios (UP)~\cite{bib75} establish the theoretical foundations of online portfolio selection by emphasizing continuous rebalancing and nonparametric universal learning, while methods such as UBAH~\cite{bib73} and the Markov of order zero (M0) model~\cite{bib74} further extend these paradigms. To capture inter-asset correlations and dynamic market characteristics, algorithms such as the Exponential Gradient (EG)~\cite{bib76} are developed, enhancing adaptability and resilience. Alternatively, a family of mean-reversion approaches, including Passive Aggressive Mean Reversion (PAMR)~\cite{bib78} and Confidence Weighted Mean Reversion (CWMR)~\cite{bib79}, exploits the mean-reverting tendencies of asset prices to stabilize predictive performance. Hybrid techniques that integrate nonparametric statistics with log-optimal portfolio theory, such as the Correlation-driven Nonparametric Learning Strategy (CORN)~\cite{bib84}, provide flexible mechanisms for leveraging structural dependencies among assets. 

Beyond these algorithmic frameworks, the classical Capital Asset Pricing Model (CAPM)~\cite{bib85} remains a cornerstone of quantitative finance, offering a rigorous theoretical basis for asset pricing and investment allocation. Despite their foundational role, traditional strategies exhibit inherent limitations, as they are largely rule-based and struggle to adapt to complex, high-dimensional, and evolving financial environments \cite{bib129}. These constraints motivate the development of machine learning and deep reinforcement learning methods, which aim to overcome such rigidity and provide more adaptive and data-driven solutions for portfolio optimization.

In addition to these classical approaches, several alternative methodologies extend portfolio optimization beyond rule-based strategies. For example, the study on stock portfolio selection using Dempster--Shafer evidence theory~\cite{bib113} applies evidence-theoretic reasoning to quantify uncertainty and integrate heterogeneous signals, while fuzzy cross-entropy, mean, variance, and skewness models~\cite{bib114} utilize fuzzy mathematics and higher-order moments to capture non-linear risk characteristics. A hybrid two-stage robustness approach~\cite{bib115} further formulates portfolio construction as a robust optimization problem under uncertainty, explicitly emphasizing resilience against model misspecification and market shocks. Beyond portfolio allocation, hybrid predictive models such as the efficient hybrid approach for forecasting real-time stock market indices~\cite{bib116} and the MMGAN-HPA algorithm for stock price prediction~\cite{bib117} leverage deep generative models and multi-model integration to improve the forecasting accuracy of asset prices. While these methods enrich the methodological landscape of financial modeling, they remain primarily optimization- or prediction-driven, with limited adaptability to high-dimensional and non-stationary portfolio management environments \cite{bib130}.

Building on these developments, reinforcement learning (RL) becomes increasingly prominent in portfolio management, with research primarily emphasizing uncertainty modeling, transfer learning, expert priors, neural architecture design, and multi-objective learning, rather than asset structuring or hierarchical decision-making. Kang et al.\ propose the Neural Process Continuous Reinforcement Learning (NPCRL)~\cite{bib94}, which integrates Neural Processes with continuous-action RL to capture market uncertainty and enable dynamic rebalancing in highly volatile environments, achieving a favorable trade-off between return and risk. Representation Transfer Reinforcement Learning (RTRL)~\cite{bib95} addresses limited samples and non-stationarity by migrating features across domains, although its benefits mainly derive from transferability rather than explicit modeling of asset structures. Choi and Kim~\cite{bib96} design an expert-infused DRL framework that incorporates tutor policies as priors to guide dynamic allocation, demonstrating that RL agents can outperform expert strategies but focusing on knowledge infusion rather than structural modeling. An early contribution by Yang~\cite{bib97} demonstrates the feasibility of end-to-end deep RL for finance by mapping price sequences directly to portfolio weights with Convolutional Neural Networks (CNNs)~\cite{bib109} and Long Short-Term Memory networks (LSTMs)~\cite{bib110}, yet it does not incorporate classification, clustering, or hierarchical structures. Recent advances explore more expressive architectures: DRL-UTrans~\cite{bib98} combines Transformers and U-Net to capture long-range dependencies and multi-scale patterns in financial time series, while Deep Long Short-Term Memory Q-Learning (DLQL) and its attention-based variant DLAQL~\cite{bib99} integrate LSTMs and attention within a Q-learning framework for applications in the oil and gas sector. Other research extends RL to multi-objective optimization, such as the framework in~\cite{bib100}, which balances return maximization and risk control through reward design, broadening optimization objectives without modeling hierarchical or structural dependencies. Collectively, these studies provide diverse perspectives on portfolio management, yet they share common limitations \cite{bib97,bib21,bib107}: insufficient treatment of structural relationships among assets, the absence of explicit classification or clustering mechanisms, and limited adaptability and interpretability in complex and volatile financial environments \cite{bib131}. These gaps highlight the need for structural innovations that move beyond network design alone and motivate the hierarchical and clustering-based approach developed in this study.

Beyond conventional portfolio optimization and generic reinforcement learning approaches, a number of studies are more directly aligned with this work. These contributions incorporate asset structuring into portfolio management through mechanisms such as dynamic subset selection, clustering, ranking, or hierarchical organization, thereby following the paradigm of structuring assets prior to portfolio decision-making. For clarity, these studies are organized into three categories, their respective limitations are examined, and the ways in which the proposed model addresses these shortcomings are demonstrated.
\begin{table}[htbp]
\centering
\caption{Comparison of closely related studies with our proposed model}
\label{tab:related}
\footnotesize
\begin{tabularx}{\linewidth}{
  >{\RaggedRight\arraybackslash}p{2.2cm}
  >{\RaggedRight\arraybackslash}X
  >{\RaggedRight\arraybackslash}X
  >{\RaggedRight\arraybackslash}X}
\toprule
\textbf{Study} & \textbf{Innovation} & \textbf{Limitation} & \textbf{Our Advantage} \\
\midrule
Dynamic Coreset Construction~\cite{bib101} &
Constructs representative subsets to approximate market distribution, improving online portfolio efficiency. &
Emphasizes sampling efficiency but ignores explicit asset correlations. &
Integrates dynamic clustering into RL with explicit cross-asset modeling. \\
\midrule
Clustering + Mean-Variance + RL~\cite{bib102} &
Combines clustering, mean-variance optimization, and RL to reflect a “classification then decision” paradigm. &
Relies on static clusters and restrictive mean-variance assumptions. &
Employs adaptive clustering and hierarchical agents, avoiding rigid constraints. \\
\midrule
CAD: Clustering + Deep RL~\cite{bib21} &
Incorporates clustering with deep RL for multi-period portfolio management. &
Treats clustering as pre-processing with no dynamic RL interaction. &
Embeds clustering directly in the RL loop with hierarchical modeling. \\
\midrule
Clustering-based PSO-CNN+MVF~\cite{bib112} &
Integrates clustering with PSO-CNN return prediction and MVF-based allocation. &
Static clustering and modular design; lacks RL-driven adaptability. &
Dynamic clustering within hierarchical RL ensures adaptive structuring. \\
\midrule
Stock Ranking and Matching RL~\cite{bib104} &
Uses ranking and matching signals to group assets for RL-based allocation. &
Heavily depends on external signals with weak structure modeling. &
Direct clustering enables adaptive and interpretable asset grouping. \\
\midrule
ASA: Autonomous Stock Selection and Allocation~\cite{bib105} &
Applies graph/hypergraph ranking for selection and regression for allocation. &
Follows a supervised paradigm with simplified risk handling. &
Dynamic clustering within RL allows interactive adaptation to shifts. \\
\midrule
HRT: Hierarchical Reinforced Trader~\cite{bib106} &
Designs a two-level hierarchy with selection at the top and execution at the bottom. &
Focuses on task partitioning rather than structural asset modeling. &
Our hierarchy captures both classification and cross-asset relationships. \\
\bottomrule
\end{tabularx}
\end{table}

In the category of clustering- or subset-based reinforcement learning methods, Dynamic Coreset Construction~\cite{bib101} dynamically constructs representative subsets to approximate the overall market distribution, thereby improving the efficiency of online portfolio selection. This mechanism can be regarded as a special form of dynamic clustering. The framework combining K-means~\cite{bib43}, mean-variance optimization, and reinforcement learning~\cite{bib102} explicitly embodies the idea of ``classification followed by decision.'' The CAD framework~\cite{bib21} further extracts asset correlations via clustering and integrates them with deep reinforcement learning for multi-period portfolio management. In a related direction, the clustering-based return prediction model with PSO-CNN+MVF~\cite{bib112} employs clustering for stock pre-selection, followed by deep return prediction and mean–variance forecasting, thereby reducing noise and enhancing the reliability of portfolio construction. These studies collectively contribute by enhancing portfolio decisions with clustering or grouping mechanisms. However, their clustering processes are mostly static, lacking dynamic interaction with reinforcement learning, which limits adaptability under non-stationary markets. In contrast, our hierarchical multi-agent framework embeds dynamic clustering directly into the training process and assigns distinct roles to lower- and upper-level agents to capture temporal dependencies and cross-asset relationships, thereby achieving stronger adaptability and structural modeling capability.

In the category of ranking- or matching-based reinforcement learning methods, the Stock Ranking and Matching RL approach~\cite{bib104} implicitly forms asset groups through ranking and matching mechanisms, and then optimizes portfolio selection via reinforcement learning. The ASA framework~\cite{bib105} employs graph and hypergraph ranking models for stock selection, combined with classification and regression models for weight allocation. This covers the entire process from selection to allocation. These methods share the contribution of introducing ranking or graph structures into portfolio management. Their limitations, however, lie in their reliance on external ranking signals or supervised learning paradigms, which reduces their dynamic adaptability and interactive learning capability, while risk constraints and transaction costs are simplified. In contrast, our method avoids reliance on external signals, directly models asset correlations via dynamic clustering, and achieves interactive and adaptive optimization within a reinforcement learning framework, making it better suited to non-stationary market conditions.  

In the category of hierarchical structures, the Hierarchical Reinforced Trader (HRT)~\cite{bib106} proposes an explicit two-level framework, where the upper level selects assets and the lower level executes transactions, thus introducing hierarchical mechanisms into portfolio management. Its contribution lies in expanding the decision structure of traditional RL. However, its hierarchy mainly targets task partitioning (selection vs.\ execution) rather than asset-structural partitioning, and therefore remains limited in capturing cross-asset relationships. By contrast, our hierarchical design is tailored for asset classification and cross-asset relationship modeling, with explicit division of roles between levels. This ensures that the hierarchical objectives are tightly coupled with market structure, enabling more effective operation under non-stationary environments.  

To more intuitively demonstrate the current research results on extracting potential connections between assets, we summarize the relevant studies and present the results in Table~\ref{tab:related}.

\section{Definition}\label{Def}

This section includes definitions of financial markets and portfolios, and relevant mathematical definitions related to market transactions. It provides the theoretical basis for our work.

\subsection{Financial Market}\label{FM}

The financial market represents a complex ecosystem in which a wide range of instruments -- such as equities, fixed income assets, and certificates of deposit -- are continuously traded with the aim of generation of returns. This study concentrates on the U.S. equity market, which is distinguished by its comprehensive regulatory framework and substantial liquidity, rendering it one of the leading markets worldwide in terms of capitalization and the scale of investor participation \cite{bib40}.

\subsection{Portfolio}\label{Portfolio}
Portfolio management constitutes a systematic investment process aimed at maximizing returns through the optimization of asset allocation. Consider an investor selecting $m$ stocks, where $m>0$, with the portfolio adjust closing price vector at time $t$ represented by the $m\times 1$ column vector
\begin{equation*}
\boldsymbol{v}_{t-1}=\left[ v_{1,t-1},v_{2,t-1}, \ldots ,v_{m-1,t-1},v_{m,t-1} \right] ^T.
\end{equation*}

For analytical purposes, the investment horizon is partitioned into $F$ discrete intervals, each associated with a period $t \in \mathbb{N}$. A period $t$ begins immediately after $t$ and concludes at $t+1$, corresponding to the open–closed interval $(t,\ t+1]$. The relative price vector of the portfolio, denoted as $\boldsymbol{z}_t$ and having dimension $1\times m$, is defined as a function of $\boldsymbol{v}_{t}^\prime$ (the portfolio price vector at the beginning of period $t$) and $\boldsymbol{v}_{t}$ (the price vector at the end of period $t$):
\begin{equation}
\boldsymbol{z}_t = \boldsymbol{v}_{t}\oslash\boldsymbol{v}_{t-1},
\nonumber
\end{equation}
where $\oslash$ indicates element-wise division. In view of arguments suggesting that upward price movements should not be regarded as a source of risk \cite{bib42}, this study utilizes the asset Sortino ratio $\boldsymbol{z}_{Sortino}$ \cite{bib42}, a $1\times m$ vector, to quantify asset risk. The Sortino ratio incorporates downside standard deviation, reflecting the perspective that only adverse volatility represents true risk \cite{bib42}. Unlike the Sharpe ratio, which penalizes both upside and downside volatility, the Sortino ratio concentrates exclusively on downside deviation, thereby providing a risk measure that better aligns with investor preferences and the asymmetric nature of financial return distributions \cite{bib42}. This property makes it particularly suitable for portfolio optimization and for integration into reinforcement learning frameworks, where distinguishing harmful volatility from favorable gains is critical for designing effective risk-aware allocation strategies \cite{bib42}. Formally, the asset Sortino ratio $\boldsymbol{z}_{\text{Sortino}}$ measures the excess return per unit of downside risk and is defined as
\begin{equation}
\boldsymbol{z}_{Sortino}=\,\,\frac{\mathbb{E}\left[ \log _2\left(\boldsymbol{z}_t\right)-\,\,r_A \right]}{\sqrt{\frac{1}{j}\sum_{\log _2\left(\boldsymbol{z}_t\right)<r_A}^j{\left( \log _2\left(\boldsymbol{z}_t\right)-\,\,r_A \right)}^2}},
\nonumber
\end{equation}
where $r_A$ denotes the minimum acceptable logarithmic return with base 2, specified as the daily logarithmic risk-free rate $r_{A}$ in this study. The $\mathbb{E}[.]$ denotes the expectation. The index $j$ corresponds to the number of periods during which the daily logarithmic return of an asset falls below this threshold. For $\boldsymbol{z}_{\text{Sortino}}$, larger values indicate stronger potential for asset appreciation.

For mitigating decision complexity, capturing structural heterogeneity, and adapting to evolving market conditions, this work applies K-means clustering \cite{bib43} for dynamic asset classification. K-means is an unsupervised learning algorithm that partitions the dataset into $k$ disjoint clusters by minimizing intra-cluster variance:
\begin{equation}
\underset{C}{\arg\min}
\sum_{i=1}^{k}
\sum_{\log_{2}(\boldsymbol{v}_{j,t-1}) \in C_i}
\big\|
\log_{2}(\boldsymbol{v}_{j,t-1}) -
\log_{2}(\boldsymbol{v}_{a_i,t-1})
\big\|_1^2,
\nonumber
\end{equation}
where $\boldsymbol{v}_{a_i,t-1}$ denotes the adjusted closing price vector of asset $a_i$, 
which serves as the representative (centroid) of cluster $C_i$, and $\left\| \cdot \right\|_1$ is the $L_1$ norm, measuring the sum of absolute changes in portfolio weights and thus quantifying the total volume of reallocation. The parameter $k$ is predetermined; in this study $k=2$, corresponding to two asset groups. As noted by Fama and French \cite{bib65}, excessive stratification may fragment portfolios and lead to unstable estimates, which further justifies our parsimonious two-group design. The algorithm iteratively alternates between assigning data points to the nearest centroid and recalculating centroids as the mean of the assigned points, until assignments stabilize or the objective function converges. To balance sensitivity to regime shifts with computational tractability, clustering is re-executed every 75 trading days, which approximately aligns with a quarterly cycle \cite{bib44,bib45}.  

The adoption of K-means clustering confirms three distinct benefits.  
First, it offers a straightforward and computationally efficient approach for categorizing large-scale asset sets \cite{bib46}.  
Second, by incorporating Sortino ratio features, the classification emphasizes risk-adjusted returns rather than raw price movements \cite{bib43}.  
Third, periodic re-execution enables dynamic restructuring of asset clusters in response to market evolution, thereby strengthening the robustness and adaptability of the overall portfolio optimization process \cite{bib47}.

\begin{figure}[htbp]
    \centering
    \includegraphics[width=10cm]{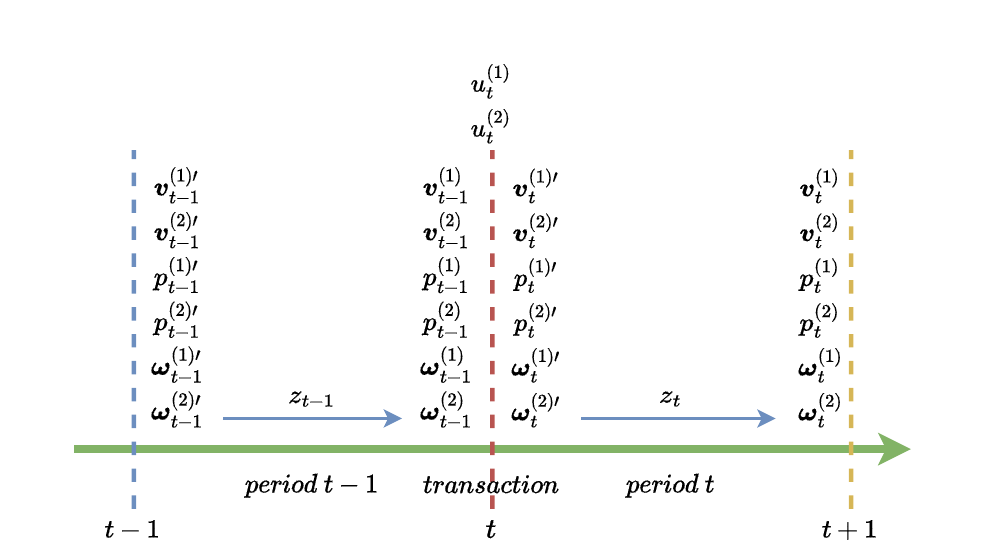}
    \caption[Transaction flow.]{
    \textbf{Transaction flow.}At the end of time $t-1$ (equivalently, at the beginning of trading period $t$), the system state is described by the price vectors $\boldsymbol{v}^{(1)}_{t-1}$ and $\boldsymbol{v}^{(2)}_{t-1}$, the group values $p^{(1)}_{t-1}$ and $p^{(2)}_{t-1}$, and the allocation weights $\boldsymbol{\omega}^{(1)}_{t-1}$ and $\boldsymbol{\omega}^{(2)}_{t-1}$. The proportional adjustments in group values induced by the rebalancing decision at time $t$ are denoted by $u^{(1)}_t, u^{(2)}_t \in [0,1]$. Immediately after rebalancing, the updated state variables become $\boldsymbol{v}^{(1)\prime}_{t}$, $\boldsymbol{v}^{(2)\prime}_{t}$, $p^{(1)\prime}_{t}$, $p^{(2)\prime}_{t}$, $\boldsymbol{\omega}^{(1)\prime}_{t}$, and $\boldsymbol{\omega}^{(2)\prime}_{t}$, which remain unchanged throughout period $t$. At the end of period $t$, the realized market state is expressed as $\boldsymbol{v}^{(1)}_{t}$, $\boldsymbol{v}^{(2)}_{t}$, $p^{(1)}_{t}$, $p^{(2)}_{t}$, $\boldsymbol{\omega}^{(1)}_{t}$, and $\boldsymbol{\omega}^{(2)}_{t}$. The relative change in asset prices during period $t$ is captured by the relative price vector $\boldsymbol{z}_t$.}
    \label{Trans}
    \end{figure}

After applying K-means clustering, the result is expressed as $C = \{C^{(1)}, C^{(2)}\}$, which partitions the asset universe into two mutually exclusive subsets. To embed this clustering outcome within the hierarchical reinforcement learning framework, binary masks $m^{(1)}_{i,t}$ and $m^{(2)}_{i,t}$ are introduced for each cluster, defined as
\begin{equation*}
\begin{aligned}
\boldsymbol{m}^{(1)}_{i,t-1} &=
\begin{cases}
1, & \text{if asset } i \in \boldsymbol{C}^{(1)}_{i,t-1} \\
0, & \text{otherwise}
\end{cases}
\\
\boldsymbol{m}^{(2)}_{i,t-1} &=
\begin{cases}
1, & \text{if asset } i \in \boldsymbol{C}^{(2)}_{i,t-1} \\
0, & \text{otherwise}
\end{cases}
\end{aligned}
,\quad
\end{equation*}
where $\boldsymbol{m}^{(1)}_{t}$ and $\boldsymbol{m}^{(2)}_{t}$ denote complementary masks of dimension $1\times m$, ensuring that only the assets associated with their respective clusters remain active. These masks are then employed to guide the lower-level agents, thereby constraining intra-group allocation to assets within each dynamically identified group. Using $\boldsymbol{m}^{(1)}_{i,t-1}$, $\boldsymbol{m}^{(2)}_{i,t-1}$, and the price vector $\boldsymbol{v}_{t-1}$, the masked price vectors $\boldsymbol{v}^{(1)}_{t-1}$ and $\boldsymbol{v}^{(2)}_{t-1}$ corresponding to Group 1 and Group 2 are obtained as
\begin{equation*}
\begin{aligned}
\boldsymbol{v}^{(1)}_{t-1} &= \boldsymbol{v}_{t-1} \odot \boldsymbol{m}^{(1)}_{t-1} \\
\boldsymbol{v}^{(2)}_{t-1} &= \boldsymbol{v}_{t-1} \odot \boldsymbol{m}^{(2)}_{t-1}
\end{aligned}
.\quad
\end{equation*}
where $\odot$ denotes the \textit{Hadamard product} \cite{bib51}. The portfolio price history over the most recent $n$ periods is represented by the matrix $\boldsymbol{q}_{t-1}$, an $m\times n$ matrix defined as
\begin{equation*}
\boldsymbol{q}_{t-1}=\left[ \boldsymbol{v}_{t-n},\boldsymbol{v}_{t-n+1}, \ldots ,\boldsymbol{v}_{t-2},\boldsymbol{v}_{t-1} \right].
\end{equation*}
Analogously, the price matrices $\boldsymbol{q}^{(1)}_{t-1}$ and $\boldsymbol{q}^{(2)}_{t-1}$ corresponding to Groups 1 and 2 are given by
\begin{equation*}
\begin{aligned}
\boldsymbol{q}^{(1)}_{t-1} &= \boldsymbol{q}_{t-1} \odot \boldsymbol{m}^{(1)}_{t-1} \\
\boldsymbol{q}^{(2)}_{t-1} &= \boldsymbol{q}_{t-1} \odot \boldsymbol{m}^{(2)}_{t-1}
\end{aligned}
.\quad
\end{equation*}
Based on the stock price matrix $\boldsymbol{q}^{(1)}_{t-1}$, $\boldsymbol{q}^{(2)}_{t-1}$ and the mask $\boldsymbol{m}^{(1)}_{t-1}$, $\boldsymbol{m}^{(2)}_{t-1}$, we obtain the logarithmic relative price matrix $\boldsymbol{lzq}^{(1)}_{t-1}$, $\boldsymbol{lzq}^{(2)}_{t-1}$:
\begin{equation*}
\begin{aligned}
\boldsymbol{lzq}^{(1)}_{t-1} = \log _2\left(\boldsymbol{q}^{(1)}_{t-1}\oslash\boldsymbol{q}^{(1)}_{t-2}\right) \\
\boldsymbol{lzq}^{(2)}_{t-1} = \log _2\left(\boldsymbol{q}^{(2)}_{t-1}\oslash\boldsymbol{q}^{(2)}_{t-2}\right)
\end{aligned}
.\quad
\end{equation*}
Within the portfolio, the allocation of capital at time $t$ is represented by the weight vector $\boldsymbol{\omega}_{t-1}$, defined as
\begin{equation*}
\boldsymbol{\omega }_{t-1}=\left[ \omega _{1,t-1},\omega _{2,t-1}, \ldots ,\omega _{m-1,t-1},\omega _{m,t-1} \right] ^T,
\end{equation*}
which is an $m \times 1$ vector. The initial weight vector is specified as the zero vector of dimension $m \times 1$, indicating that the initial endowment is entirely allocated to cash. At any time $t$, the portfolio weights satisfy the budget constraint
\begin{equation*}
\sum_{i=1}^m{\omega }_{i,t-1}=1.
\end{equation*}
Correspondingly, the group-specific allocations are denoted by $\boldsymbol{\omega}^{(1)}_{t-1}$ and $\boldsymbol{\omega}^{(2)}_{t-1}$, defined through the masking mechanism:
\begin{equation*}
\begin{aligned}
\boldsymbol{\omega}^{(1)}_{t-1} &= \boldsymbol{\omega}_{t-1} \odot \boldsymbol{m}^{(1)}_{t-1} \\
\boldsymbol{\omega}^{(2)}_{t-1} &= \boldsymbol{\omega}_{t-1} \odot \boldsymbol{m}^{(2)}_{t-1} \\
1 &= \sum_{j=1}^m{\omega}^{(1)}_{j,t-1} + \sum_{i=1}^m{\omega}^{(2)}_{j,t-1} + \omega^{(f)}_{t-1}
\end{aligned}
,\quad
\end{equation*}
where $\omega^{(f)}_{t-1}$ is the weight of risk free asset at time $t$. The portfolio value prior to trading at time $t$, denoted as $p_{t-1}$, is determined jointly with the share vector $\boldsymbol{sh}_{t-1}$, which represents the number of shares hold across assets. These quantities are obtained as
\begin{equation*}
\begin{aligned}
\boldsymbol{sh}_{t-1} &= \boldsymbol{sh}^{(1)}_{t-1} + \boldsymbol{sh}^{(2)}_{t-1}, \\
p^{(1)}_{t-1}             &= \boldsymbol{sh}^{(1)T}_{t-1} \boldsymbol{v}^{(1)}_{t-1} + h^{(1)}_{t-1} \\
\boldsymbol{sh}^{(1)}_{t-1} 
                     &= \left\lfloor 
                        p^{(1)\prime}_{t-1} \, \boldsymbol{\omega}^{(1)\prime}_{t-1} 
                        \oslash \boldsymbol{v}^{(1)\prime}_{t-1} 
                        \right\rfloor \\
p^{(2)}_{t-1}             &= \boldsymbol{sh}^{(2)T}_{t-1} \boldsymbol{v}^{(2)}_{t-1} + h^{(2)}_{t-1} \\
\boldsymbol{sh}^{(2)}_{t-1} 
                     &= \left\lfloor 
                        p^{(2)\prime}_{t-1} \, \boldsymbol{\omega}^{(2)\prime}_{t-1} 
                        \oslash \boldsymbol{v}^{(2)\prime}_{t-1} 
                        \right\rfloor
\\
p^{(f)}_{t-1}             &=p_{t-1}^\prime \omega^{(f)\prime}_{t-1} 2^{r_A} \\
p_{t-1}              &= p^{(1)}_{t-1} + p^{(2)}_{t-1} + p^{(f)}_{t-1} 
\end{aligned}
,\quad
\end{equation*}
where, $\boldsymbol{sh}^{(1)}_{t-1}$, $\boldsymbol{sh}^{(2)}_{t-1}$, $h^{(1)}_{t-1}$, and $h^{(2)}_{t-1}$ denote the share holdings and residual cash for Groups 1 and 2, respectively, at time $t$. Portfolio values of Group 1, 2 and risk free asset at time $t$ are denoted as $p^{(1)}_{t-1}$, $p^{(2)}_{t-1}$ and $pf_{t-1}$. Similarly, $p^{(1)\prime}_{t-1}$, $p^{(2)\prime}_{t-1}$, $pf_{t-1}^\prime$ and $\boldsymbol{\omega}^{(1)\prime}_{t-1}$, $\boldsymbol{\omega}^{(2)\prime}_{t-1}$, $\omega^{(f)\prime}_{t-1}$ denote portfolio values and weight vector of Groups 1, 2 and risk free asset at beginning of period $t-1$. The operator $\oslash$ specifies element-wise division, while the floor operator $\lfloor \cdot \rfloor$ enforces integer rounding of the number of shares by truncation.
Refer to Figure~\ref{Trans} for a comprehensive illustration of the transaction mechanism. At the end of period $t-1$ (i.e., at time $t$), the price vectors, portfolio values, and allocation weights of the two groups are denoted by $\boldsymbol{v}^{(1)}_{t-1}$, $\boldsymbol{v}^{(2)}_{t-1}$, $p^{(1)}_{t-1}$, $p^{(2)}_{t-1}$, $\boldsymbol{\omega}^{(1)}_{t-1}$, and $\boldsymbol{\omega}^{(2)}_{t-1}$, respectively. Once transactions at time $t$ are executed, the updated states of the system are represented as $\boldsymbol{v}^{(1)\prime}_{t}$, $\boldsymbol{v}^{(2)\prime}_{t}$, $p^{(1)\prime}_{t}$, $p^{(2)\prime}_{t}$, $\boldsymbol{\omega}^{(1)\prime}_{t}$, and $\boldsymbol{\omega}^{(2)\prime}_{t}$. The portfolio is then maintained without additional rebalancing until the conclusion of period $t$.  

Each transaction occurs instantaneously at the decision point. Upon completion, the group-specific price vectors $\boldsymbol{v}^{(1)}_{t-1}$ and $\boldsymbol{v}^{(2)}_{t-1}$ are updated to $\boldsymbol{v}^{(1)\prime}_{t}$ and $\boldsymbol{v}^{(2)\prime}_{t}$, the group values $p^{(1)}_{t-1}$ and $p^{(2)}_{t-1}$ become $p^{(1)\prime}_{t}$ and $p^{(2)\prime}_{t}$, and the weight vectors $\boldsymbol{\omega}^{(1)}_{t-1}$ and $\boldsymbol{\omega}^{(2)}_{t-1}$ transform into $\boldsymbol{\omega}^{(1)\prime}_{t}$ and $\boldsymbol{\omega}^{(2)\prime}_{t}$, all reflecting the newly determined investment strategy. For analytical tractability, this study assumes that trading activity does not influence asset prices, i.e., $\boldsymbol{v}^{(1)}_{t-1}=\boldsymbol{v}^{(1)\prime}_{t}$ and $\boldsymbol{v}^{(2)}_{t-1}=\boldsymbol{v}^{(2)\prime}_{t}$ \cite{bib7}. Consequently, the price vectors at the start of period $t$ are represented as $\boldsymbol{v}^{(1)\prime}_{t}$ and $\boldsymbol{v}^{(2)\prime}_{t}$. The initial group values $p^{(1)\prime}_{t}$ and $p^{(2)\prime}_{t}$ at the beginning of period $t$ are expressed as
\begin{equation*}
\begin{aligned}
p^{(1)\prime}_{t} &= u^{(1)}_t \, p^{(1)}_{t-1} \\
p^{(2)\prime}_{t} &= u^{(2)}_t \, p^{(2)}_{t-1}
\end{aligned}
,\quad
\end{equation*}
where $u^{(1)}_t$ and $u^{(2)}_t$ denote the proportional reductions in group values at time $t$ due to transaction costs.

Transaction costs are deducted from group values, denoted by $Cost^{(1)}_t$ and $Cost^{(2)}_t$. The fee structure is formalized as:
\begin{equation*}
\begin{aligned}
Cost^{(1)}_t &= cs \, p^{(1)}_{t-1} \left\| \boldsymbol{\omega}^{(1)}_{t-1} - \boldsymbol{\omega}^{(1)\prime}_t \right\|_1 \\
Cost^{(2)}_t &= cs \, p^{(2)}_{t-1} \left\| \boldsymbol{\omega}^{(2)}_{t-1} - \boldsymbol{\omega}^{(2)\prime}_t \right\|_1
\end{aligned}
,\quad
\end{equation*}
where $cs$ represents the transaction cost rate. Following Jiang et al.~\cite{bib7}, the transaction cost rate $c$ is fixed at 0.001.
After rebalancing, the residual cash holdings $h^{(1)}_t$ and $h^{(2)}_t$ that remain unallocated to equities are expressed as
\begin{equation*}
\begin{aligned}
h^{(1)}_t &= \left\{ \begin{array}{l}
	\frac{(h^{(1)}_{t-1} + h^{(2)}_{t-1}) \cdot \sum_{i=1}^m{\omega}^{(1)}_{i,t-1}}{\sum_{i=1}^m{\omega}^{(1)}_{i,t-1} + \sum_{i=1}^m{\omega}^{(2)}_{i,t-1}}-p^{(1)\prime}_{t}+p^{(1)}_{t-1}, \quad \text{if } t \ge 2 \\
	\frac{p_0 \cdot \sum_{i=1}^m{\omega}^{(1)}_{i,t-1}}{\sum_{i=1}^m{\omega}^{(1)}_{i,t-1} + \sum_{i=1}^m{\omega}^{(2)}_{i,t-1}}-p^{(1)\prime}_{t}+p^{(1)}_{t-1}, \quad \text{if } t=1
\end{array} \right.
\\
h^{(2)}_t &= \left\{ \begin{array}{l}
	\frac{(h^{(1)}_{t-1} + h^{(2)}_{t-1}) \cdot \sum_{i=1}^m{\omega}^{(2)}_{i,t-1}}{\sum_{i=1}^m{\omega}^{(1)}_{i,t-1} + \sum_{i=1}^m{\omega}^{(2)}_{i,t-1}}-p^{(2)\prime}_{t}+p^{(2)}_{t-1}, \quad \text{if } t \ge 2 \\
	\frac{p_0 \cdot \sum_{i=1}^m{\omega}^{(2)}_{i,t-1}}{\sum_{i=1}^m{\omega}^{(1)}_{i,t-1} + \sum_{i=1}^m{\omega}^{(2)}_{i,t-1}}-p^{(2)\prime}_{t}+p^{(2)}_{t-1}, \quad \text{if } t=1
\end{array} \right.
\end{aligned}
,\quad
\end{equation*}
where $p_0$ denotes the initial portfolio value. If any allocation leads to $h^{(1)}_t < 0$ or $h^{(2)}_t < 0$, the transaction is proportionally scaled down to guarantee non-negativity of the cash balance, i.e., $h^{(1)}_t, h^{(2)}_t \geq 0$. This mechanism prohibits debt accumulation and enforces that all asset positions are backed by available capital. The same feasibility constraint is consistently applied across all settings to ensure a fair and comparable evaluation.  

The value of the portfolio at the beginning of period $t$, denoted $p_t^\prime$, is given by
\begin{equation*}
\begin{aligned}
p^{(1)\prime}_t = u^{(1)}_t p^{(1)}_{t-1} &= \left\{
\begin{array}{l}
\frac{(h^{(1)}_{t-1} + h^{(2)}_{t-1}) \cdot \sum_{i=1}^m{\omega}^{(1)}_{i,t-1}}{\sum_{i=1}^m{\omega}^{(1)}_{i,t-1} + \sum_{i=1}^m{\omega}^{(2)}_{i,t-1}} - h^{(1)}_t + p^{(1)}_{t-1}, \quad \text{if } t \ge 2 \\
\frac{p_0 \cdot \sum_{i=1}^m{\omega}^{(1)}_{i,t-1}}{\sum_{i=1}^m{\omega}^{(1)}_{i,t-1} + \sum_{i=1}^m{\omega}^{(2)}_{i,t-1}} - h^{(1)}_t + p^{(1)}_{t-1}, \quad \text{if } t = 1
\end{array}
\right.
\\
p^{(2)\prime}_t = u^{(2)}_t p^{(2)}_{t-1} &= \left\{
\begin{array}{l}
\frac{(h^{(1)}_{t-1} + h^{(2)}_{t-1}) \cdot \sum_{i=1}^m{\omega}^{(2)}_{i,t-1}}{\sum_{i=1}^m{\omega}^{(1)}_{i,t-1} + \sum_{i=1}^m{\omega}^{(2)}_{i,t-1}} - h^{(2)}_t + p^{(2)}_{t-1}, \quad \text{if } t \ge 2 \\
\frac{p_0 \cdot \sum_{i=1}^m{\omega}^{(2)}_{i,t-1}}{\sum_{i=1}^m{\omega}^{(1)}_{i,t-1} + \sum_{i=1}^m{\omega}^{(2)}_{i,t-1}} - h^{(2)}_t + p^{(2)}_{t-1}, \quad \text{if } t = 1
\end{array}
\right.
\end{aligned}
.\quad
\end{equation*}
The transaction return in period $t$ is quantified through the base-2 logarithmic rate of return $\varphi_t$, defined as
\begin{equation*}
\begin{aligned}
\varphi_t   &= \log _2\left(\frac{p_{t}}{p_{t-1}}\right) \\
\varphi^{(1)}_t  &= \log _2\left(\frac{p^{(1)}_{t}}{p^{(1)}_{t-1}}\right) \\
\varphi^{(2)}_t  &= \log _2\left(\frac{p^{(2)}_{t}}{p^{(2)}_{t-1}}\right)
\end{aligned}
,\quad
\end{equation*}
where $\varphi^{(1)}_t$ and $\varphi^{(2)}_t$ denote group-specific returns. This iterative process continues for all periods, and the final portfolio value $p_f$ is expressed as
\begin{equation}
p_f = p_0 \cdot 2^{ \sum_{t=1}^{F} \varphi_t },
\nonumber
\end{equation}
with $F$ indicating the total number of investment periods and $\varphi_0$ is the initial capital. In this study, the initial capital is set to 1 million dollars.

\subsection{Assumptions}\label{Assumption}

The analysis is based on two foundational assumptions, denoted as A1 and A2.

A1: The market possesses sufficient liquidity to enable the prompt execution of each transaction.

A2: Transactions are assumed to have no impact on stock prices.

According to Assumption A1, if a stock exhibits sufficiently high liquidity such that its trading volume exceeds the required execution size, all proposed trades can be executed without delay. Assumption A1 is essential to the framework, as daily market volume cannot be forecasted with precision. In practical markets, large transaction volumes can influence stock prices because both buying and selling activities convey investors’ sentiments and are reflected in price fluctuations. Therefore, Assumption A2 is introduced to abstract from this price-impact effect and maintain model tractability. Taken together, these two assumptions hold reasonably well when the selected stocks exhibit sufficiently high liquidity~\cite{bib7}.

Although a transaction cost of 0.1$\%$ per transaction is incorporated in the experimental design to simulate basic market frictions \cite{bib7}, several critical aspects of real-world trading—such as slippage, order book depth constraints, and market impact—remain insufficiently modeled \cite{bib123}. Under such conditions, applying the proposed strategy in live trading environments may result in execution price deviations \cite{bib123} and partial fills \cite{bib124}, which significantly deteriorate the realized returns compared to backtests results. Furthermore, the multi-asset coordinated rebalancing paths learned by the agent during training may become infeasible due to real-world execution constraints, thereby impairing portfolio optimization effectiveness and overall strategy stability. Nevertheless, the proposed framework is equipped to jointly capture both long-term trends and short-term market dynamics, enabling it to adaptively modulate trading frequency based on prevailing market regimes. This multi-timescale perception mechanism offers greater flexibility and robustness compared to strategies relying solely on single-horizon signals, contributing to improved resilience against execution errors and performance degradation induced by market frictions \cite{bib41}. It is worth noting that, although these idealized assumptions may lead to systematic overestimation of the absolute performance across all evaluated strategies, each approach—ranging from traditional rule-based baselines to alternative reinforcement learning frameworks—is assessed under the same experimental conditions. Therefore, the simplification applies uniformly across methods, ensuring the relative fairness of performance comparisons and preserving the validity of the observed inter-strategy rankings.

\section{Framework of DRL}\label{Framework}
\subsection{Overall Framework}
\begin{figure}[htbp]
    \centering
    \includegraphics[width=12cm,height=6.5cm]{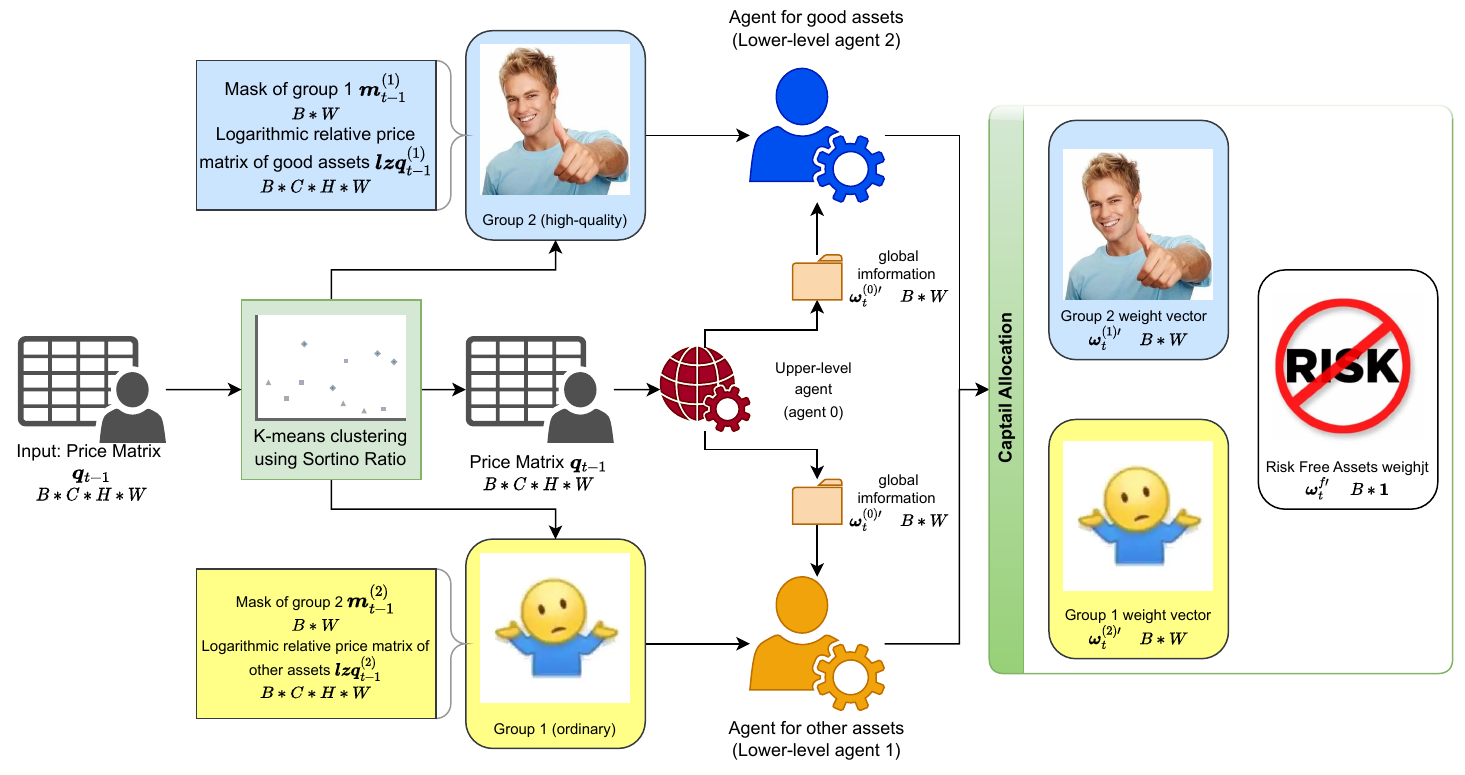}
    \caption[Workflow of the proposed architecture.]{
    \textbf{Workflow of the proposed architecture.}The state representation is first processed through dynamic clustering and masking, after which the upper-level agent extracts global signals and the lower-level agents perform intra-group allocation. A fusion mechanism then integrates the outputs to produce the final portfolio weights, ensuring that alongside concentrated investment in high-quality assets, the inclusion of remaining assets provides diversification benefits that reduce overall portfolio risk \cite{bib85}.}
    \label{Workflow}
\end{figure}

The proposed structure combines the continuous-control capability of the Deep Deterministic Policy Gradient (DDPG) \cite{bib49} algorithm with the task-decomposition advantage of HRL \cite{bib48}. The upper-level agent provides stable global representations and guidance, while the lower-level agents perform fine-grained allocation within dynamically defined asset groups. Through this design, the framework establishes coherent linkages between global and local perspectives, thereby enhancing feature utilization, allocation granularity, and interpretability. The overall workflow of the proposed architecture is illustrated in Figure~\ref{Workflow}.

\subsubsection{Hierarchical Reinforcement Learning Design}
Hierarchical reinforcement learning decomposes and coordinates complex tasks through the cooperation of upper- and lower-level agents \cite{bib48}. In the proposed design, the upper-level agent (Agent 0) extracts global market signals and is then kept fixed to provide a stable supervisory interface, while the lower-level agents (Agent 1 and Agent 2) operate under dynamic masking constraints to allocate weights across ordinary and high-quality asset groups. This structure enables simultaneous modeling of global and local decision information, thereby enhancing representational richness and refining allocation granularity.

\subsubsection{Hierarchical Decision Mechanism}
Leveraging dynamic asset grouping and allocation process is explicitly integrated into a unified hierarchical reinforcement learning framework. Unlike rule-based subset updates \cite{bib21,bib101,bib102,bib112} or implicit black-box selections \cite{bib7,bib97}, the mechanism enforces explicit information flow and cross-layer consistency constraints. This design improves interpretability of the decision process and ensures coherent coordination between global and local decision modules.

\subsection{State, Action and Reward}
\begin{figure}[htbp]
    \centering
    \includegraphics[width=8cm,height=4cm]{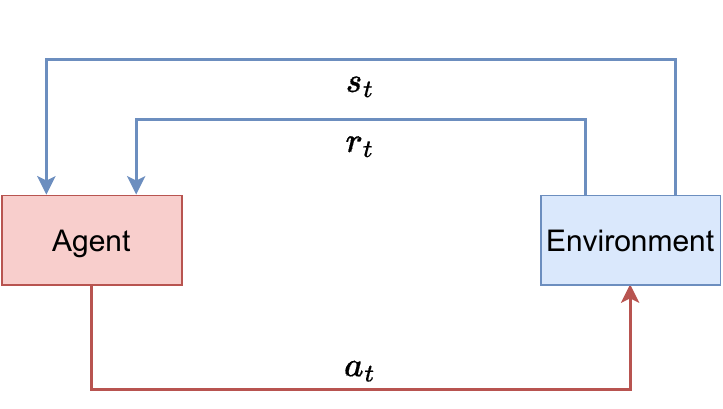}
    \caption[Closed-loop interaction.]{
    \textbf[Closed-loop interaction.]Closed-loop interaction between agent and environment in reinforcement learning, where the agent observes the state $s_t$, takes action $a_t$, and receives the reward $r_t$ together with the next state $s_{t+1}$.}
    \label{agent_and_environment}
\end{figure}

In reinforcement learning (RL), the interaction between the agent and the environment follows a closed-loop process, as illustrated in Figure~\ref{agent_and_environment}. At each decision step $t$, the agent observes the environment through the state $s_t$, generates an action $a_t$ based on its policy, and applies this action to the environment. The environment then responds by updating the system dynamics and returning both a new state $s_{t+1}$ and a scalar reward $r_{t+1}$. This iterative process allows the agent to optimize its policy so as to maximize cumulative rewards over time \cite{bib50}. 
In the context of portfolio optimization, the environment corresponds to the financial market, the state encapsulates observable market information and portfolio status, the action represents allocation decisions across assets, and the reward reflects investment performance \cite{bib7}. The following subsections elaborate on the detailed design of state, action, and reward in this study.

In reinforcement learning, the state $s_t$ refers to the complete information observed by the agent from the environment at time $t$, which provides the basis for decision-making. As illustrated in the standard agent–environment interaction loop, the environment updates the portfolio value after receiving the action $a_t$ and returns the new state $s_{t+1}$ together with the immediate reward $r_{t+1}$. The construction of the state thus determines the agent's perception of the market environment \cite{bib3}.

In the portfolio optimization problem, the environment corresponds to the financial market, and the state should capture market dynamics, portfolio positions, and structural constraints. In this study, the state at time $t$ is defined as  
\begin{equation*}
s_t = \{ \boldsymbol{q}_{t-1}, \boldsymbol{sh}^{(1)}_{t-1}, \boldsymbol{sh}^{(2)}_{t-1}, \boldsymbol{m}^{(1)}_{t-1}, \boldsymbol{m}^{(2)}_{t-1} \}
,\quad
\end{equation*}
where
\begin{itemize}
    \item $\boldsymbol{q}_{t-1} \in \mathbb{R}^{m \times n}$: the price history matrix of the past $n$ days, reflecting the temporal evolution of the market;
    \item $\boldsymbol{sh}^{(1)}_{t-1}, \boldsymbol{sh}^{(2)}_{t-1}$: the portfolio positions of the two asset groups \emph{before the current transaction}, ensuring that decisions are conditioned on existing holdings rather than reinitialized at each step;
    \item $\boldsymbol{m}^{(1)}_{t-1}, \boldsymbol{m}^{(2)}_{t-1} \in \{0,1\}^m$: binary masks generated by dynamic clustering, indicating the tradable subsets at time $t$.
\end{itemize}

This state representation is shared across all three agents -- one upper-level agent and two lower-level agents---thereby providing a unified market perception. Such a shared-state design ensures consistent information flow throughout the hierarchical framework, while allowing each agent to concentrate on its designated decision-making responsibility \cite{bib125}. By jointly encoding temporal information, path-dependent holdings, and dynamically evolving market structures, this formulation enhances adaptability and robustness in non-stationary financial environments.

In reinforcement learning, the action $a_t$ denotes the agent's policy output under state $s_t$, which directly influences portfolio evolution and determines future returns. In the portfolio optimization setting, the action corresponds to the allocation weights assigned to assets. At time $t$, the actions $a^{(0)}_t$ and $a^{(1)}_t$ is defined as
\begin{equation*}
\begin{aligned}
a^{(0)}_t = \{\boldsymbol{\omega}^{(1)\prime}_t\} \\
a^{(1)}_t = \{\boldsymbol{\omega}^{(2)\prime}_t\}
\end{aligned}
.\quad
\end{equation*}
Here, $\omega1_t^\prime$ is nonzero only for assets indicated by $m1_t$, and $\omega2_t^\prime$ is nonzero only for assets indicated by $m2_t$. Together, they satisfy the budget constraint:
\begin{equation*}
\sum_{j=1}^m \omega^{(1)\prime}_{j,t} + \sum_{j=1}^m \omega^{(2)\prime}_{j,t} + \omega^{f\prime}_t = 1.
\end{equation*}
This formulation decomposes the global allocation problem into two structured subproblems, allowing the agent to optimize allocations within each group while maintaining overall capital conservation. 

In reinforcement learning, the reward $R_t$ provides the immediate feedback signal that drives policy optimization \cite{bib50}. In the proposed hierarchical framework, the asset universe is dynamically partitioned into two subsets through K-means clustering \cite{bib43}, where the clustering is periodically executed on rolling windows of asset returns to capture regime shifts and structural changes in the market. For each group $i \in \{1,2\}$, the instantaneous reward $R^{(i)}_t$ is based on the logarithmic return $\log_2 \!\left( \tfrac{p^{(i)}_t}{p^{(i)}_{t-1}} \right)$. To stabilize the learning process, this base reward $R^{(i)}_t$ is further defined as
\begin{equation*}
R^{(i)}_t =
\begin{cases}
\kappa \cdot \log_2 \left( \dfrac{p^{(i)}_t}{p^{(i)}_{t-1}} \right), & N_{t-1} < 2, \\[1.0em]
\kappa \cdot \Bigg( \log_2 \left( \dfrac{p^{(i)}_t}{p^{(i)}_{t-1}} \right) 
     - \beta^{adj}_{t-1} \cdot \sigma^{(i)}_{t-1} \Bigg), & N_{t-1} \geq 2,
\end{cases}
\end{equation*}
where $\kappa$ is a positive scaling coefficient introduced to amplify the reward signal, ensuring sufficient gradient magnitude during training and stabilizing policy optimization \cite{bib3}. The term $\sigma^{(i)}_{t-1}$ denotes the standard deviation of historical log-returns for group~$i$ at time $t$, and $N_{t-1}$ represents the number of past observations available at time $t$. The adaptive risk-aversion coefficient $\beta^{adj}_{t-1}$ is then defined as
\begin{equation*}
\beta^{adj}_{t-1} = \beta \cdot \frac{N_{t-1}}{\eta},
\end{equation*}
where $\beta$ is a fixed hyperparameter set to $0.2$ in this study and $\eta$ is a normalization constant representing the baseline number of observations \cite{bib19}. Here, $\beta$ serves as the baseline risk-aversion factor that regulates the strength of volatility penalization in the reward function: a smaller $\beta$ prioritizes return maximization with limited regard for risk, whereas a larger $\beta$ enforces more conservative behavior by amplifying the influence of volatility \cite{bib126}. Multiplication by $\tfrac{N_{t-1}}{\eta}$ allows the overall coefficient $\beta^{adj}_{t-1}$ to increase as the number of observations grows, thereby reducing the weight of risk penalties when estimates are noisy in early stages and strengthening robustness as more reliable information becomes available \cite{bib126}.

The group-wise reward formulation establishes a one-to-one correspondence between each reward signal and the agent responsible for its designated asset group. Specifically, $R^{(1)}_t$ is assigned to agent~1 managing asset group~1, while $R^{(2)}_t$ is assigned to agent~2 managing asset group~2. This design ensures that the optimization objectives remain consistent with the structural decomposition of the market, while simultaneously enhancing interpretability, since the contribution of each asset group to policy learning can be explicitly identified.

\subsection{Clustering and Mask Mechanism}

To introduce structured constraints into the hierarchical reinforcement learning framework, this study applies the K-means algorithm \cite{bib43} to partition assets into two groups: \textbf{Group~1 (high-quality assets)} and \textbf{Group~2 (ordinary assets)}. For each group, a binary mask vector is constructed to restrict the input space of the lower-level agents. For the asset universe at time $t$, the clustering result is denoted as $C_{t-1} = \{C1_{t-1}, C2_{t-1}\}$. The corresponding masks are defined as
\begin{equation*}
\begin{aligned}
\boldsymbol{m}^{(1)}_{j,t-1} &=
\begin{cases}
1, & \text{if asset $j$ belongs to Group~1 (high-quality)} \\
0, & \text{otherwise}
\end{cases} \\
\boldsymbol{m}^{(2)}_{j,t-1} &=
\begin{cases}
1, & \text{if asset $j$ belongs to Group~2 (ordinary)} \\
0, & \text{otherwise}
\end{cases}
\end{aligned}
.\quad
\end{equation*}
At any time $t$, the masks are complementary, satisfying $\boldsymbol{m}^{(1)}_{t-1} + \boldsymbol{m}^{(2)}_{t-1} = \mathbf{1}$. Based on these masks, the subset input logarithmic return matrix matrices $\boldsymbol{z}^{(1)}_{t-1}$, $\boldsymbol{z}^{(2)}_{t-1}$ are obtained:
\begin{equation*}
\begin{aligned}
\boldsymbol{z}^{(1)}_{t-1} &= \boldsymbol{z}_{t-1} \odot \boldsymbol{m}^{(1)}_{t-1} \\
\boldsymbol{z}^{(2)}_{t-1} &= \boldsymbol{z}_{t-1} \odot \boldsymbol{m}^{(2)}_{t-1} \\
\boldsymbol{z}_{t-1} &= \log_2 \left(\boldsymbol{q}_{t-1} \oslash \boldsymbol{q}_{t-2}\right)
\end{aligned}
.\quad
\end{equation*}
Correspondingly, the lower-level agents output the group-specific weight vectors $\boldsymbol{\omega}^{(1)}_{t-1}$ and $\boldsymbol{\omega}^{(2)}_{t-1}$, which allocate capital within Group~1 and Group~2, respectively:
\begin{equation*}
\begin{aligned}
\boldsymbol{\omega}^{(1)\prime}_{t} &= \boldsymbol{\omega^\prime_{t}} \odot \boldsymbol{m}^{(1)}_{t-1} \\
\boldsymbol{\omega}^{(2)\prime}_{t} &= \boldsymbol{\omega^\prime_{t}} \odot \boldsymbol{m}^{(2)}_{t-1}
\end{aligned}
.\quad
\end{equation*}
Through this clustering and mask mechanism, the model conducts group-wise learning, ensuring that each lower-level agent operates strictly within its designated subset of assets. This design reduces the complexity of the action space and establishes a clear structural constraint for subsequent weight fusion and global decision-making \cite{bib4,bib37}.

\subsection{DDPG Framework}
DDPG is a canonical actor–critic based deep reinforcement learning approach that is suitable for control tasks in continuous action spaces \cite{bib49}. In this framework, the actor network outputs deterministic actions, such as portfolio weights, while the critic network evaluates state–action pairs and updates parameters through policy gradients. Because DDPG directly models continuous numerical outputs, it aligns with portfolio optimization, where weight adjustments are inherently continuous.

\subsection{Actor--Critic Network Structure}

The upper-level agent employs a transformer-based architecture validated in prior financial applications~\cite{bib15}. This choice underscores that our contribution lies in structural innovation, while relying on established networks for function approximation. This structure employs a transformer-based model that effectively captures temporal dependencies in financial time series and cross-sectional correlations among assets~\cite{bib121}. Compared with conventional recurrent or convolutional networks, the transformer-based architecture exhibits stronger capability in extracting global market features. The spatial interaction and hierarchical relationship between the upper-level and lower-level agents are illustrated in Figure~\ref{Workflow}. As depicted, clustered and masked states are processed by the upper-level and lower-level agents in parallel, and a fusion module integrates their outputs into the final portfolio weights, preserving both focus and diversification \cite{bib85}.
\begin{figure}[htbp]
    \centering
    \includegraphics[width=6cm]{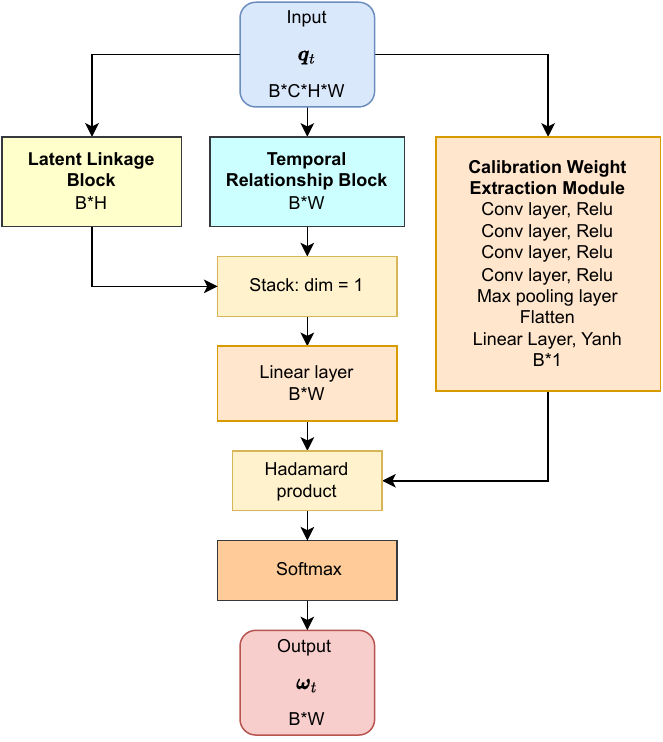}
    \caption[Overall workflow of the proposed framework.]{
    \textbf[Overall workflow of the proposed framework.]Overall workflow of the proposed hierarchical reinforcement learning framework~\cite{bib15}. The framework consists of an upper-level agent, two lower-level agents, and an environment module. The upper-level agent captures global market information and generates coordination signals, which guide the lower-level agents in asset-specific allocation. Each lower-level agent focuses on a distinct subset of assets, extracting local temporal patterns and making portfolio decisions accordingly. The environment provides market observations and reward feedback to complete the learning loop. Detailed descriptions of the module composition and implementation are available in the original work~\cite{bib15}.}
    \label{upper_level}
\end{figure}
As illustrated in Fig.~\ref{upper_level}, the hierarchical reinforcement learning framework comprises an upper-level agent and two lower-level agents that interact through structured information exchange. It is particularly noteworthy that the upper-level agent directly receives the complete price information $\boldsymbol{q}_{t-1}$, enabling it to capture inter-asset correlation structures and price dynamics. Its implicit output, denoted as $\boldsymbol{\omega}^{(0)\prime}_{t}$, serves as a global guidance signal that encapsulates this information and is subsequently provided to the lower-level agents for coherent coordination. In contrast, the lower-level agents operate on logarithmic return matrices under mask constraints $\boldsymbol{z}^{(1)}_{t-1}$, $\boldsymbol{z}^{(2)}_{t-1}$, masks $\boldsymbol{m}^{(1)}_{t-1}$, $\boldsymbol{m}^{(2)}_{t-1}$, and information $\boldsymbol{\omega}^{(0)\prime}_{t}$ obtained from the upper-level agent. The focus of the lower-level agents is confined to intra-group weight allocation, which ensures coherence between global objectives and local optimization processes. As a result, they lack the capacity to represent global cross-asset relationships. Through this hierarchical design, the upper-level agent supplements the information unavailable to the lower-level agents and provides guidance signals for more coherent decision making.

\begin{figure}[htbp]
    \centering
    \includegraphics[width=12cm,height=8cm]{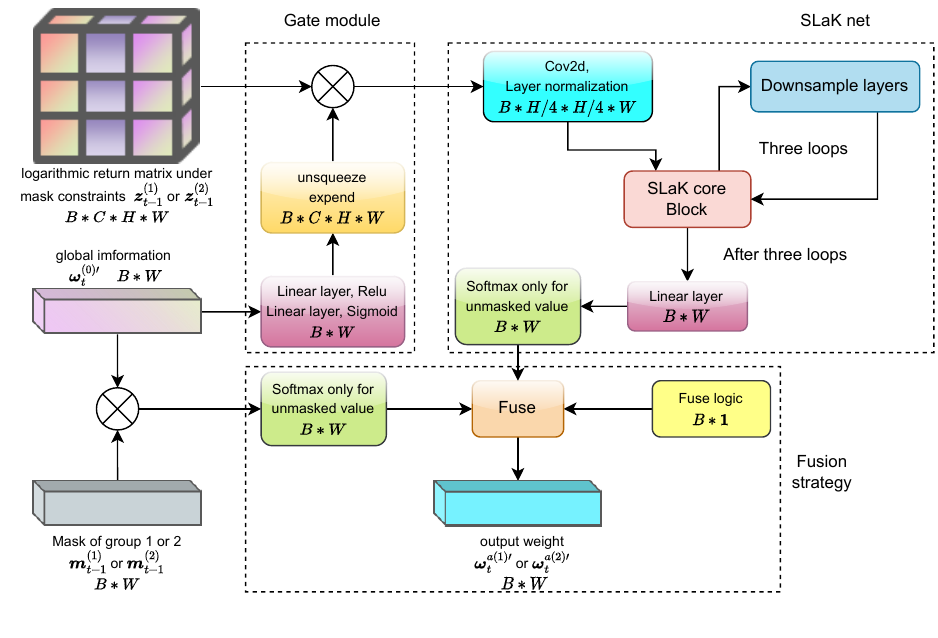}
    \caption[Architecture of the lower-level actor network.]{
    \textbf[Architecture of the lower-level actor network.]Architecture of the lower-level actor network. The network integrates masked logarithmic return matrices, global information from the upper-level agent, and a SLaK backbone to generate intra-group weight distributions, which are subsequently fused with prior signals to obtain group-wise portfolio weights.}
    \label{lower-level actor network}
\end{figure}

As shown in Figure~\ref{lower-level actor network}, lower-level agents~1 and~2 receive as input the masked logarithmic return matrices $\boldsymbol{q}^{(1)}_{t-1}$ and $\boldsymbol{q}^{(2)}_{t-1}$, corresponding to the high-quality asset group and the ordinary asset group, respectively. Both agents share an identical network architecture, with differences arising solely from the masks $\boldsymbol{m}^{(1)}_{t-1}$ and $\boldsymbol{m}^{(2)}_{t-1}$, which define the effective input domain by filtering out irrelevant assets and thereby reducing the dimensionality of the allocation space \cite{bib122}. 
The network consists of three main components:  

1. \textbf{Global information integration (Gate module).}  
The global signal $\boldsymbol{\omega}^{(0)\prime}_{t}$ produced by the upper-level agent is first transformed through nonlinear mappings to generate gating coefficients. These coefficients perform element-wise multiplication with the input matrices, allowing the lower-level agents to incorporate global market guidance in their intra-group allocation and avoid relying solely on local patterns.

2. \textbf{Sparse Large Kernel Network (SLaK) backbone.}  
The gated return matrix is processed by a SLaK convolutional backbone adopted from \cite{bib52}. SLaK employs sparsity-enabled kernels with receptive fields extending beyond $51 \times 51$, allowing the network to jointly model temporal dependencies and intra-group correlations. This design substantially enhances representational capacity while preserving computational tractability. The resulting logits are linearly projected and combined with the group masks $\boldsymbol{m}^{(1)}_{t-1}$ and $\boldsymbol{m}^{(2)}_{t-1}$ to filter valid entries, followed by a softmax operation that produces the intra-group weight distributions $\boldsymbol{\omega}^{(1)\prime}_{t}$ and $\boldsymbol{\omega}^{(2)\prime}_{t}$ 

3. \textbf{Fusion strategy within lower-level agents.}  
The softmax-based intra-group weights are further combined with the upper-level guidance signal. Specifically, the masked and normalized output of the upper-level agent is computed as  
\begin{equation*}
\tilde{\boldsymbol{\omega}^{(0)\prime}_{t}} = \mathrm{softmax}\big(\boldsymbol{\omega}^{(0)\prime}_{t} \odot \boldsymbol{m}_i^{t-1}\big), \quad i \in \{1,2\},
\end{equation*}
which represents a group-wise prior distribution. This prior is integrated with the intra-group predictions $\boldsymbol{\omega}^{(i)\prime}_{t}$, where $i \in \{1,2\}$, through a convex combination \cite{bib121}, that is, a linear combination with non-negative coefficients summing to one. The combination ratio is controlled by a learnable coefficient $\lambda$, constrained to $(0,1)$ via a sigmoid function, thereby enabling adaptive adjustment between reliance on prior knowledge and data-driven predictions. The fused intra-group allocation is expressed as:  
\begin{equation*}
\boldsymbol{\omega}^{a(i)\prime}_{t} = (1-\lambda)\,\tilde{\boldsymbol{\omega}^{(0)\prime}_{t}} + \lambda\,\boldsymbol{\omega}^{(i)\prime}_{t}, \quad i \in \{1,2\}
.\quad
\end{equation*}
This design provides several benefits:  
\begin{itemize}
    \item \textbf{Consistency of information:} The fusion mechanism ensures coherent coordination between global and local decision modules \cite{bib54}.  
    \item \textbf{Finer allocation granularity:} The design preserves the detailed representation power of intra-group modeling while embedding global constraints \cite{bib53}.  
    \item \textbf{Enhanced interpretability:} The final outputs can be decomposed into global guidance and local optimization, offering transparent insights into the decision process \cite{bib50}.  
\end{itemize}  
It should be emphasized that the outputs $\boldsymbol{\omega}^{(1)\prime}_{t}$ and $\boldsymbol{\omega}^{(2)\prime}_{t}$ represent only relative intra-group weights. The final portfolio weights are determined in the subsequent capital allocation and cross-group fusion mechanism.

The lower-level critic network adopts a transformer-based architecture for value estimation as introduced in \cite{bib15}. In this setting, the critic encodes state–action pairs and learns the corresponding value function, thereby providing optimization signals for actor updates. The principal modification lies in the incorporation of masking constraints at the input stage. Specifically, for each group $i \in \{1,2\}$, the state $\boldsymbol{z}^{(i)}_{t-1}$ and the action $\boldsymbol{\omega}^{(i)\prime}_{t}$ are element-wise multiplied with the corresponding mask $\boldsymbol{m}^{(i)}_{t-1}$ before being fed into the critic, ensuring that value estimation remains confined to the designated asset subset. This design prevents cross-group interference and preserves group-specific independence. Two advantages arise from this modification. First, it maintains coherence with the hierarchical framework, in which the upper-level agent extracts global signals while the lower-level critic performs localized value assessment \cite{bib54}. Second, it enforces consistency between value estimation and dynamic classification, thereby reinforcing the logical closure of the hierarchical decision process \cite{bib53}.

\subsection{Capital Allocation and Final Portfolio Weights via Utility Optimization}
After obtaining the intra-agent weights $\boldsymbol{\omega}^{a(1)\prime}_{t}$ and $\boldsymbol{\omega}^{a(2)\prime}_{t}$, we integrate them with classical portfolio selection techniques to determine the allocation of investment capital among the risk-free asset, Group~1, and Group~2. Within the domain of portfolio theory, the mean–variance framework represents one of the most influential and widely applied methodologies. Subsequent developments extend this paradigm by formulating portfolio selection as the maximization of an expected utility function, a perspective that generalizes the trade-off between risk and return. When the utility function is quadratic, the problem of maximizing expected utility can be equivalently expressed as minimizing portfolio variance while maximizing expected return, as discussed in \cite{Bodnar-Parola-Schmid-a}.  

In this formulation, the investor seeks to optimize the expected utility of terminal wealth, with preferences encoded through a chosen utility function \cite{Canakoulu_and_Ozekici_portfoil_exponential}. Standard measures of risk aversion are well established in the literature \cite{Pratt_Risk_aversion,Arrow_risk_bearing}, providing formal foundations for modeling investor behavior. Moreover, \cite{Bertsekas_dynamic_programming} analyzes a class of utility functions and derives optimal multi-period policies under dynamic settings. Leveraging these theoretical foundations, section 4.6 formulates an exponential utility framework to represent investor preferences and derive optimal allocations among the risk-free asset, Group~1, and Group~2.

\subsubsection{Maximization of expected utility function}
Prior to addressing the primary problem, it is essential to establish some notational conventions. At the commencement of period \( t \), an investor possessing an endowment represented by \( W_t \) is tasked with determining the optimal portfolio weights \( \boldsymbol{\omega^{\prime}_t} \) to allocate towards risky assets. This allocation is aimed at maximizing the expected utility of their wealth in the subsequent period, while investing the remainder, represented by \( 1 - \sum \boldsymbol{\omega^{\prime}_t} \), in a risk-free asset. The wealth function of these assets can be articulated as follows:

\[
W_{t} = W_{t-1} (1 + r_{A}) + W_{t-1}\left[ \boldsymbol{\omega^{\prime\top}_t} (R_t - \mathbf{1}r_{A}) \right].
\] 

This formulation encompasses both the risk-free rate and the adjustments for risky asset returns, thereby facilitating a comprehensive analysis of wealth dynamics in the investment strategy. Therefore, the portfolio selection problem is 
\begin{equation*}\label{selection_problem}
    \max_{\boldsymbol{\omega^{\prime}_t}} \mathbb{E}\left[  U(W_{t})\right].
\end{equation*}

In many instances, obtaining an analytical solution to the expected utility maximization problem is notably challenging, prompting numerous scholars to depend on numerical methods, as referenced in \cite{Brandt_numerical} and \cite{Canakoulu_and_Ozekici_portfoil_exponential}. Nevertheless, a significant number of researchers have successfully derived analytical solutions for various scenarios of the expected utility maximization problem, particularly when the utility function is exponential, as indicated in \cite{Bodnar_Close_Solution} and \cite{Rasonyi2025exponential}. 

In the subsequent section, we will articulate the primary contributions derived from our research. In this study, we do not investigate the relationship between optimal portfolio weights and wealth. Consequently, we set the wealth level at \( W_t = 1 \). Our utility function is expressed as follows: 
\begin{equation*}\label{utility_function}
    U(x) = -e^{-\alpha x} \quad \alpha >0. 
\end{equation*}

Assuming that the log-return conforms to a Normal distribution, it follows that the expression \( R_t - \mathbf{1} r_{A} \) also follows to a Normal distribution. As a result, the wealth function can be redefined as follows: 

\begin{equation*}
W_t \overset{d}{=} W_{t-1} (1 + r_{A}) + W_{t-1} \left[ \boldsymbol{\omega}^{\prime\top}_t (\boldsymbol{\mu} - \boldsymbol{\mathbf{1}} r_{A}) + \sqrt{\boldsymbol{\omega}^{\prime\top}_t \boldsymbol{\Sigma} \boldsymbol{\omega}^{\prime}_t} N_{(0, 1)}\right],
\end{equation*}
where \(\mu\) represents the expected value of log-returns (at the period $t$), \(\Sigma\) denotes the covariance of log-returns (at the period $t$), and \(N_{(0, 1)}\) is interpreted as the standard normal distribution. Therefore, the expectation of the exponential utility function can be derived as follows:

\begin{equation*}
\begin{split}
    \mathbb{E}\left[ U(W_{t}) \right] &= - e^{-\alpha W_{t-1}(1 + r_{A})} e^{-\alpha W_{t-1}\left[ \boldsymbol{\omega}^{\prime\top}_t(\boldsymbol{\mu} - \mathbf{1}r_{A}) \right]} \\
    &\times \int_{-\infty}^{+\infty} e^{ -\alpha W_{t-1}\sqrt{\boldsymbol{\omega}^{\prime\top}_t \boldsymbol{\Sigma} \boldsymbol{\omega}^{\prime}_t} x} \varphi(x)dx,
\end{split}
\end{equation*}  
where \( \varphi(x)\) denotes the density function of the standard normal distribution.

\subsubsection{Optimal solution of maximization}
To reformulate the expected exponential utility function, we shall utilize the Fourier transform, as illustrated in the following expression:

\begin{equation*}
    \max_{\boldsymbol{\omega}^{\prime}_t} -e^{-\alpha W_{t-1} (1 + r_{A})} e^{\frac{\alpha^2W_{t-1}^2}{2} \boldsymbol{\omega}^{\prime\top}_t \boldsymbol{\Sigma} \boldsymbol{\omega}^{\prime}_t - \alpha W_{t-1}\left[ \boldsymbol{\omega}^{\prime\top}_t(\boldsymbol{\mu} - \mathbf{1}r_{A}) \right]}.
\end{equation*} 

Given that \(\alpha > 0\), the constant term in the aforementioned objective function is negative, which indicates that the function is decreasing. Therefore, this optimization problem can be reformulated as a quadratic programming problem:

\begin{equation*}\label{Qoptt}
\begin{split}
 \min_{\boldsymbol{\omega^{\prime}_t \in \mathbb{R}^d}} \; \;  & \frac{\alpha^2 W_{t-1}^2}{2} \boldsymbol{\omega}^{\prime\top}_t \boldsymbol{\Sigma} \boldsymbol{\omega}^{\prime}_t, \\
s.t.\;\;  & \boldsymbol{\omega}^{\prime\top}_t(\boldsymbol{\mu} - \boldsymbol{\mathbf{1}}r_{A}) = -c.
\end{split}
\end{equation*}

It is important to note that the initial problem presents significant challenges for analytical resolution. Furthermore, when the dimensionality of the stock variables is exceedingly high, obtaining a numerical solution becomes exceptionally arduous. In this context, we define a hyperplane characterized by the equation \(\boldsymbol{\omega}^{\prime\top}_t(\boldsymbol{\mu} - \boldsymbol{\mathbf{1}}r_{A}) = -c\). Achieving the optimal solution situated below this hyperplane demonstrates enhanced efficiency. Additionally, it is feasible to modify the value of \(c\) in response to fluctuating market conditions. This adjustment enables the repositioning of the hyperplane, facilitating the identification of the optimal solution in relation to it.

The solution of this problem can be derived using the Lagrangian method \cite{JinXuZhou2008}. Accordingly, the Lagrangian function is defined as follows:

\begin{equation*}\label{Lagrangian_function}
    \mathcal{L}(\boldsymbol{\omega}_{t}) := \frac{\alpha^2 W_{t-1}^2}{2} \boldsymbol{\omega}^{\prime\top}_t \boldsymbol{\Sigma} \boldsymbol{\omega}^{\prime}_t - \lambda \Big(\boldsymbol{\omega}^{\prime\top}_t(\boldsymbol{\mu} - \mathbf{1}r_{A}) + c\Big),
\end{equation*}
where \(\lambda\) denotes the Lagrangian multiplier. By applying the first-order conditions (FOC), one arrives at a system of equations articulated as follows:
\begin{equation*}\label{sys_equs}
\Bigg\{
    \begin{aligned}
        \frac{\partial \mathcal{L}}{\partial \boldsymbol{\omega}^{\prime}_{t}} &= \alpha^2 \boldsymbol{W_{t-1}^2  \Sigma \omega_t} - \lambda (\boldsymbol{\mu} - \boldsymbol{\mathbf{1}}r_{A}) = 0\\ 
        \frac{\partial \mathcal{L}}{ \partial \lambda} &= -\boldsymbol{\omega}^{\prime\top}_t(\boldsymbol{\mu} - \boldsymbol{\mathbf{1}}r_{A}) - c = 0.
    \end{aligned}
\end{equation*}

In analyzing the system of equations for a specified constant \(c\), we denote the resultant solution as \(\boldsymbol{\omega^{\prime c}_t}\), which can be expressed in the following manner:

\[
\boldsymbol{\omega^{\prime c}_t} := -\frac{c}{B} \odot \boldsymbol{A}.
\]
where \(B = \boldsymbol{(\mu} - \boldsymbol{\mathbf{1}} r_{A})^{\top} \boldsymbol{\Sigma}^{-1} (\boldsymbol{\mu} - \boldsymbol{\mathbf{1}} r_{A})\) and \(\boldsymbol{A} = \boldsymbol{\Sigma}^{-1} (\boldsymbol{\mu} - \boldsymbol{\mathbf{1}} r_{A})\). From the analysis presented above, it can be observed that an increase in the parameter \( c \) results in a decrease in the allocation to risky assets, denoted as \( \boldsymbol{\omega}^{\prime c}_t \). This indicates that \( c \) serves as a lever to manage investment exposure to risk. Consequently, in a bearish market environment, it is advisable to decrease the value of \( c \). Conversely, during a bullish market, increasing the value of \( c \) would be advisable.

\subsubsection{Momentum-based Adjustment}
To capture the momentum of the recent market, we implement a momentum strength parameter \cite{bib103}. Initially, we utilize the log-returns from three distinct periods as the foundation for assessing momentum strength. We define momentum strength as the \( L^2 \) norm of the mean log-returns for these periods. Consequently, we introduce a new parameter denoted as \( \beta \), represented mathematically as follows:

\begin{equation*}
    \beta = Clip(3\bar{R}_{t-3:t}, l, h),
\end{equation*}
where \( \bar{R} \) refers to the mean log-returns, $l$ is the lowest value and $h$ is the highest value. The allocation of weights to risky assets is then expressed by the equation:

\begin{equation*}
    \boldsymbol{\omega}^{\prime *}_t = \boldsymbol{\omega}^{\prime c}_t\times \left(1 + \left(\beta \odot \tanh(\bar{R}_{t-3:t})\right)\right).
\end{equation*} 

This formulation allows for a structured approach to investment strategy based on identified momentum dynamics. 

To complement this momentum-based adjustment, we further incorporate a rebound detection mechanism \cite{bib108}. The objective is to distinguish between genuine rebounds following a significant decline and mere technical recoveries that may occur during an ongoing downtrend.

To avoid allocating heavily during temporary technical recoveries in ongoing downtrends, we incorporate a \textbf{rebound detection mechanism}. Let $\bar{R}_{t-m:t}$ denote the recent mean log-return for a group of assets. A rebound is identified if two conditions are simultaneously satisfied:

\begin{enumerate}
    \item \textbf{Downtrend condition:} the mean return over the preceding $m$ periods (e.g. $m=3$) falls below a negative threshold $\theta_d$, representing a sustained decline:
    \[
        \bar{R}_{t-m:t} < \theta_d
    \]
    \item \textbf{Rebound condition:} the most recent one or two periods exhibit returns above a positive threshold $\theta_u$:
    \[
        \bar{R}_{t-1} > \theta_u \quad \text{and} \quad \bar{R}_{t} > \theta_u
    \]
\end{enumerate}

Binary indicators of rebounds are computed for each asset group:

\[
\text{rebound\_flags} = [\text{rebound}_0, \text{rebound}_1]
\]

These flags act as validation signals for whether the portfolio should consider adjustments beyond the baseline momentum allocation. To integrate multiple sources of portfolio information while maintaining stability, we implement a \textbf{selective fusion mechanism}. 

The baseline weights \(\boldsymbol{\omega}^{\prime *}_{t}\), is integrated with the generated global masked information, denoted \(\boldsymbol{\omega}^{(0)i}_{t}\), specifically for asset groups that exhibit confirmed rebounds. Formally, for each group \(i\), the updated weights are defined as follows:

\[
\boldsymbol\omega^{new(i)\prime}_{t} = 
\begin{cases} 
(1-\eta)\,\boldsymbol\omega^{(i)\prime *}_{t} + \eta\,\boldsymbol\omega^{0(i)\prime}_{t}, & \text{if } \text{rebound\_flags}_i = \text{True} \\[1mm]
\boldsymbol\omega^{(i)\prime *}_{t}, & \text{otherwise}
\end{cases}
\]
where \(\eta \in [0,1]\) dictates the intensity of the blending process. The agent weights are calculated by selectively masking the relevant assets within each group and aggregating their individual actions:

\[
{\boldsymbol\omega}^{0(i)\prime}_{t} = \sum_j (\boldsymbol\omega^{(0)\prime}_{j,t} \cdot {m}_{j,t}^{(i)}).
\]

This targeted fusion ensures that adjustments informed by adaptive models affect asset allocation only when the market exhibits verified rebound behavior. Asset groups without rebound signals remain consistent with the baseline expected-utility maximizing portfolio, thereby maintaining robustness. To incorporate the risk-free asset into the allocation, the unnormalized logits are constructed by concatenating the fixed baseline element with the outputs of the two risky-asset groups as
\begin{equation*}
\boldsymbol{\omega}^{set\prime}_t = \big[0, \, \boldsymbol\omega^{new(1)\prime}_{t}, \, \boldsymbol\omega^{new(2)\prime}_{t} \big],
\end{equation*}
where the first element corresponds to the risk-free asset and is fixed at zero. Setting its logit to zero provides a stable reference baseline in the softmax normalization, reflecting the financial role of the risk-free asset as a benchmark with negligible sensitivity to market conditions. This treatment eliminates redundant parameterization and leverages the shift-invariance property of the softmax function \cite{bib127}. Accordingly, the logits of the two risky asset groups, $\boldsymbol\omega^{new(1)\prime}_{t}$ and $\boldsymbol\omega^{new(2)\prime}_{t}$, quantify their relative attractiveness against the risk-free baseline. After applying the softmax transformation,  
\begin{equation*}
\boldsymbol{\omega}^{capital\prime}_t = \mathrm{softmax}\!\left( \frac{\boldsymbol{\omega}^{set\prime}_t}{Tep} \right),
\end{equation*}
where ${Tep}>0$ denotes a temperature coefficient controlling the sharpness of the allocation distribution \cite{bib127}, we obtain normalized weights ${\boldsymbol\omega}^{capital(f)\prime}_t$, ${\boldsymbol\omega}^{capital(1)\prime}_t$, and ${\boldsymbol\omega}^{capital(2)\prime}_t$, which can be interpreted as probabilities of allocating capital across the risk-free asset and the two risky groups. Given the group-level capital proportions ${\boldsymbol\omega}^{capital(1)\prime}_t$ and ${\boldsymbol\omega}^{capital(2)\prime}_t$, together with the intra-group weight vectors $\boldsymbol{\omega}^{a(1)\prime}_{t}$ and $\boldsymbol{\omega}^{a(2)\prime}_{t}$, the final transaction-level allocations $\boldsymbol{\omega}^{(1)\prime}_{t}$ and $\boldsymbol{\omega}^{(2)\prime}_{t}$ are computed as  
\begin{equation*}
\boldsymbol{\omega}^{(i)\prime}_{t} = \boldsymbol{\omega}^{a(i)\prime}_{t} \, \boldsymbol{\omega}^{capital(i)\prime}_t, \quad i \in \{1,2\},
\end{equation*}
thereby integrating intra-group optimization with group-level capital allocation in a financially coherent manner. This design treats the risk-free asset as a \emph{benchmark-relative anchor} \cite{bib85}, ensuring that risky asset allocations are consistently interpreted with respect to a stable financial baseline.

\subsection{Loss Function and Exploration Strategy}

The optimization of the lower-level framework is based on the DDPG paradigm, augmented with auxiliary components and enhanced exploration mechanisms. 

\subsubsection{Critic objective.}  
The critic network is optimized using a temporal-difference (TD) objective \cite{bib55}, the canonical method in reinforcement learning for estimating value functions through bootstrapping from subsequent rewards. This formulation is derived from the Bellman equation \cite{bib55}, which establishes the recursive structure of value estimation. For a transition $(\boldsymbol{s}_{t-1}, \boldsymbol{a}_{t-1}, r_t, \boldsymbol{s}_t)$, the objective is expressed as
\begin{equation*}
\mathcal{L}_{\text{critic}}
= \mathbb{E}\Big[\,
\big( r_t + \gamma Q_{\theta^-}(\boldsymbol{s}_t, \pi_{\phi^-}(\boldsymbol{s}_t)) 
- Q_\theta(\boldsymbol{s}_{t-1}, \boldsymbol{a}_{t-1}) \big)^2
\,\Big],
\end{equation*}
where $Q_{\theta}$ represents the critic network parameterized by $\theta$, which estimates the expected return of a state–action pair. The policy $\pi_{\phi}$, parameterized by $\phi$, denotes the actor network that generates deterministic portfolio-weight actions. The pairs $(\theta^-, \phi^-)$ correspond to the delayed target parameters used to stabilize training by reducing estimation variance and preventing oscillations during learning.

\subsubsection{Actor network objective with detached prior alignment.}  
The actor network is optimized by maximizing the critic’s evaluation of its policy outputs, augmented with two auxiliary components: an entropy regularizer and a detached prior-alignment penalty. The overall objective is formulated as
\begin{equation*}
\begin{aligned}
\mathcal{L}_{\text{actor}}
&= -\,\mathbb{E}_{\boldsymbol{s}_{t}}
\Big[\, Q_\theta(\boldsymbol{s}_{t}, \pi_\phi(\boldsymbol{s}_{t})) \,\Big]
\ -\ \beta\,\mathcal{H}\!\big(\pi_\phi(\boldsymbol{s}_{t})\big)
\ +\ \alpha\,\mathcal{L}^{\text{det}}_{\text{imit}}, \\[0.6em]
\mathcal{H}\!\big(\pi_\phi(\boldsymbol{s}_{t})\big)
&= - \sum_k p_k \log p_k,\quad
\tilde{\boldsymbol{\omega}}^{(0,i)\prime}_{t}
= \mathrm{softmax}\!\big(\boldsymbol{\omega}^{(0)\prime}_{t} \odot \boldsymbol{m}^{(i)}_{t-1}\big), \\[0.6em]
\mathcal{L}^{\text{det}}_{\text{imit}}
&= \big\|\,
\operatorname{sg}\!\big(\boldsymbol{\omega}^{(i)\prime}_{t}\big)
\ -\
\operatorname{sg}\!\big(\tilde{\boldsymbol{\omega}}^{(0,i)\prime}_{t}\big)
\big\|_2^2, \quad i \in \{1,2\},
\end{aligned}
\end{equation*}
where $\mathcal{H}(\cdot)$ denotes the entropy of the action distribution, $\alpha=0.1$ and $\beta=0.01$ are fixed coefficients, and $\operatorname{sg}(\cdot)$ indicates the stop-gradient operator. Because both terms in $\mathcal{L}^{\text{det}}_{\text{imit}}$ are detached, this component introduces no gradient contributions to network parameters and therefore acts as a constant offset in optimization. Its function is to provide a stable diagnostic of the alignment between intra-group predictions $\boldsymbol{\omega}^{(i)\prime}_{t}$ and the masked upper-level prior $\tilde{\boldsymbol{\omega}}^{(0,i)\prime}_{t}$, while strictly preventing cross-layer gradient propagation \cite{bib55}.

\subsubsection{Exploration mechanism.}  
To ensure effective exploration in the continuous, high-dimensional action space, the framework employs a hybrid exploration scheme comprising:
\begin{itemize}
    \item \textbf{Ornstein–Uhlenbeck (OU) process noise.} Correlated Gaussian perturbations generated by an OU process are added to the actions, producing temporally coherent trajectories that better capture trading dynamics \cite{bib56}.
    \item \textbf{$\epsilon$-greedy perturbation.} With probability $\epsilon$, the action is perturbed by zero-mean Gaussian noise and subsequently clipped to admissible bounds, thereby injecting stochasticity and mitigating entrapment in local optima \cite{bib57}.
\end{itemize}
Together, these components balance exploration and exploitation: OU-process noise promotes stability and temporal smoothness, whereas $\epsilon$-greedy perturbations enhance resilience in volatile market conditions.

\section{Experiment}\label{Experiment}
\subsection{Data}\label{Data}

We construct three independent datasets from Yahoo Finance\footnote[1]{\url{http://www.finance.yahoo.com}}, with the Dow Jones Industrial Average (DJIA) serving as the empirical foundation of the analysis. The DJIA is selected as the benchmark for the SAMP-HDRL framework because it comprises 30 highly liquid equities from diverse industries, thereby providing a representative market environment consistent with the assumptions of this study. Each dataset spans a four-year horizon, in which the first three years function as the training set and the final year is reserved for backtest, as summarized in Table~\ref{tab:datasets}. To evaluate the robustness of SAMP-HDRL under extreme market turbulence induced by exogenous shocks, particular emphasis is placed on the backtest set of Dataset~2 and the training set of Dataset~3, especially those reflecting the pronounced volatility observed during the COVID-19 pandemic in 2020~\cite{bib58}. 

While reinforcement learning provides adaptive mechanisms for sequential decision-making, it cannot fully resolve the challenges posed by non-stationary market dynamics. Hence, before evaluating model performance, it is necessary to characterize the statistical properties of the datasets, particularly the 2020 segment where the COVID-19 pandemic triggered circuit breakers and induced structural breaks in U.S. equity markets. For this purpose, we apply two standard diagnostic tests to the DJIA. As reported in Table~\ref{tab:adf-arch}, the Augmented Dickey–Fuller (ADF) test \cite{bib59} fails to reject the unit-root null hypothesis, confirming that the price series is non-stationary, whereas the ARCH–LM test \cite{bib60} strongly rejects the null hypothesis of homoscedasticity, indicating pronounced conditional heteroscedasticity and volatility clustering in the return sequence. Taken together, these findings confirm that the 2020 market exhibits both non-stationarity and time-varying volatility, consistent with the exceptional turbulence of that year. Alongside the upward-trending regime in 2019 \cite{bib88,bib89} and the oscillatory dynamics of 2021 \cite{bib93}, these diagnostics establish a rigorous basis for evaluating SAMP-HDRL under heterogeneous environments, thereby enabling a comprehensive assessment of robustness and adaptability.
\begin{table}[htbp]
\centering
\caption{Unit root and conditional heteroscedasticity tests for DJIA in 2020}
\label{tab:adf-arch}
\small
\begin{tabular}{lcc}
\toprule
Test & Statistic & $p$-value \\
\midrule
ADF (Price series) & $-1.7258$ & $0.4179$ \\
ARCH--LM (Return series) & $99.0891$ & $<1 \times 10^{-16}$ \\
\bottomrule
\end{tabular}
\end{table}
Because Dow Inc.\ was separated from DowDuPont in March 2019, consistent historical records before March 18, 2019, are unavailable \cite{bib111}. Accordingly, this stock is excluded from the analysis, leaving 29 constituents in the sample. The trading horizon is defined on a daily scale, and the adjusted closing price of each equity is employed to construct the price matrix. Compared with raw closing prices, adjusted closing prices correct for distortions arising from dividends and stock splits~\cite{bib62}, thereby providing a more accurate, comparable, and methodologically coherent representation of asset values~\cite{bib63}.

\begin{table*}[htbp]
\begin{center}
\caption{Data Range for training datasets and backtest datasets}
\label{tab:datasets}
\resizebox{.8\textwidth}{!}{%
\begin{tabular}{lll}
\noalign{\smallskip}\hline
Data Purpose & \multicolumn{1}{c}{Training Data Range} & \multicolumn{1}{c}{Backtest Data Range}  \\ 
\noalign{\smallskip}\hline\noalign{\smallskip}
Backtest 1  & 2016/01/01 00:00 to 2019/01/01 00:00    & 2019/01/01 00:00 to 2020/01/01 00:00      \\ 
Backtest 2  & 2017/01/01 00:00 to 2020/01/01 00:00    & 2020/01/01 00:00 to 2021/01/01 00:00      \\ 
Backtest 3  & 2018/01/01 00:00 to 2021/01/01 00:00    & 2021/01/01 00:00 to 2022/01/01 00:00      \\ 
\noalign{\smallskip}\hline
\end{tabular}%
}
\end{center}
\end{table*}

\subsection{Comparison strategy}\label{C-strategy}

We benchmark our framework against nine traditional strategies and nine reinforcement learning (RL)–based approaches, three of which are obtained from the TradeMaster platform \cite{bib64}. The deep reinforcement learning (DRL) baselines include Ensemble of Identical Independent Evaluators (EIIE) \cite{bib7}, FinRL \cite{bib66}, Ensemble Stock Transactions (EST) \cite{bib67}, State-Augmented RL (SARL) \cite{bib64}, Investor-Imitator (II) \cite{bib64}, TradeMaster (PPO) \cite{bib64}, the portfolio policy network with cost sensitivity (PPN) \cite{bib68}, TARN \cite{bib69}, Deep Modern Portfolio Theory (DeepMPT) \cite{bib70}, LSRE-CAAN \cite{bib71}, and Financial Transformer Reinforcement Learning (FTRL) \cite{bib15}. This collection spans widely adopted DRL benchmarks as well as recent approaches that have gained academic visibility, ensuring both methodological breadth and relevance for comparison.  

EIIE introduces a DRL framework originally applied to cryptocurrency portfolios and demonstrates superior performance relative to heuristic strategies; this design is extended to the stock domain as a comparator against SAMP-HDRL. FinRL provides a standardized platform for automated trading, offering unified data preprocessing, environment construction, and training pipelines, thereby ensuring consistency and reproducibility. EST ensembles A2C, DDPG, and PPO to exploit their complementary strengths, representing a multi-algorithm baseline. SARL fuses heterogeneous data sources to capture high-dimensional features and improve robustness, while II encodes trading expertise within a DRL paradigm, enhancing performance across diverse metrics. TradeMaster (PPO) applies automated machine learning for adaptive hyperparameter selection within PPO. PPN emphasizes latent inter-asset dependencies and introduces a cost-sensitive reward to strengthen risk control, whereas TARN explicitly models inter-asset correlations to balance risk and return under dynamic conditions. Two recent approaches that have attracted significant academic attention since 2023 are also considered. DeepMPT integrates Modern Portfolio Theory’s risk–return principles with DRL optimization, enabling adaptive allocation in non-stationary environments, while LSRE-CAAN designs efficient representations, action spaces, and reward functions to capture temporal dependencies and market non-stationarity in high-frequency settings. Finally, FTRL employs a Transformer-based architecture to model temporal dependencies and cross-asset interactions, thereby addressing the limitations of conventional time-series models in portfolio optimization and providing a strong benchmark under comparable conditions. Detailed specifications of all baselines are summarized in Table~\ref{strategies}.  

To ensure fairness and computational consistency across all baselines, a unified hyperparameter protocol is adopted. A common learning rate is applied across frameworks to ensure consistent training dynamics, taking into account the limited training horizon (755 daily steps), the architectural complexity of Transformer-based models, and the use of automatic mixed precision (AMP) \cite{bib119}, which stabilizes optimization through reduced gradient magnitudes. Batch size remains fixed for all methods, and a uniform early-stopping criterion is enforced. Architecture-specific hyperparameters, such as the number of layers or attention heads, are not standardized, since these components are absent in certain baselines (e.g., traditional heuristics or simpler DRL models). To prevent bias from extensive hyperparameter tuning and to guarantee reproducibility, exhaustive grid search is deliberately avoided \cite{bib120}. Instead, a restricted set of widely adopted values is manually validated and consistently applied throughout the experimental setting.

\begin{table*}[htbp]
\centering
\begin{center}
\caption{The comparison strategies and their classification}
\label{strategies}
\resizebox{.8\textwidth}{!}{%
\begin{tabular}{ll}
\noalign{\smallskip}\hline
Benchmarks                                         & Uniform Constant Rebalanced Portfolios (CRP) \cite{bib72}          \\
                                                   & Uniform Buy And Hold (UBAH) \cite{bib73}                               \\
                                                   & Markov of order zero (M0)\cite{bib74}                                  \\
\noalign{\smallskip}\hline\noalign{\smallskip}
Follower-the-Winner                                & Universal Portfolios (UP) \cite{bib75}                                 \\
                                                   & Exponential Gradient (EG) \cite{bib76}                                 \\
\noalign{\smallskip}\hline\noalign{\smallskip}
Follower-the-Lose                                  & Passive Aggressive Mean Reversion (PAMR) \cite{bib78}                  \\
                                                   & Confidence Weighted Mean Reversion (CWMR) \cite{bib79}                 \\

\noalign{\smallskip}\hline\noalign{\smallskip}
Pattern-Marching                                   & Correlation-driven Nonparametric Learning Strategy (CORN) \cite{bib84} \\
\noalign{\smallskip}\hline\noalign{\smallskip}
Financial Theory                                   & Capital Asset Pricing Model (CAPM) \cite{bib85}                        \\
\noalign{\smallskip}\hline\noalign{\smallskip}
Reinforcement learning strategies                  & EIIE \cite{bib7}                                                      \\
                                                   & Finrl \cite{bib66}                                                     \\
                                                   & EST \cite{bib67}                                                       \\
                                                   & SARL \cite{bib64}                                                      \\
                                                   & II \cite{bib64}                                                        \\
                                                   & PPO \cite{bib64}                                                       \\
                                                   & PPN \cite{bib68}                                                       \\
                                                   & TARN \cite{bib69}                                                      \\
                                                   & DeepMPT \cite{bib70}                                                   \\
                                                   & LSRE-CAAN \cite{bib71}                                                 \\
                                                   & FTRL \cite{bib15}                                                      \\
\noalign{\smallskip}\hline
\end{tabular}%
}
\end{center}
\end{table*}

\subsection{Performance Measures}\label{Measures}

The evaluation of the proposed framework is conducted along two primary dimensions: profitability and risk. Profitability is measured by the final cumulative return $R_f$, defined as
\begin{equation}
R_f = \sum_{t=1}^{F} \varphi_t,
\nonumber
\end{equation}
where $\varphi_t$ denotes the portfolio return at time $t$. A larger $R_f$ reflects stronger overall profitability across the backtest horizon.  
Risk assessment employs three well-established metrics: the Sharpe ratio $r_{\text{Sharpe}}$ \cite{bib86}, the Sortino ratio $r_{\text{Sortino}}$ \cite{bib42}, and the Omega ratio $r_{\text{Omega}}$ \cite{bib87}. The Sharpe ratio quantifies excess return relative to total volatility:
\begin{equation}
r_{\text{Sharpe}} = \frac{ \mathbb{E} \!\left[ \varphi_t - r_{A} \right] }{ \sqrt{ \text{Var} \!\left( \varphi_t \right) } },
\nonumber
\end{equation}
where $\mathbb{E}$ and $\text{Var}$ denote the expectation and variance operators, and $r_{A}$ is the daily log risk-free rate.  
Recognizing that upward price fluctuations should not be classified as risk \cite{bib42}, the Sortino ratio isolates downside volatility:
\begin{equation}
r_{\text{Sortino}} = \frac{ \mathbb{E}\!\left[ \varphi_t - r_A \right] }{ \sqrt{ \tfrac{1}{j} \sum_{\varphi_t < r_A}^j \!\left( \varphi_t - r_A \right)^2 } },
\nonumber
\end{equation}
where $r_A$ is the minimum acceptable return (set to $r_{A}$ in this study), and $j$ is the number of periods with $\varphi_t < r_A$. A higher $r_{\text{Sortino}}$ reflects superior performance relative to downside risk.  
The Omega ratio evaluates the entire distribution of returns, incorporating higher-order moments:
\begin{equation}
r_{\text{Omega}} = \frac{ \sum_{\varphi_t \geq r_A}^j \!\left( \varphi_t - r_A \right) }{ \sum_{\varphi_t < r_A}^j \!\left( r_A - \varphi_t \right) },
\nonumber
\end{equation}
with $r_A$ again denoting the minimum acceptable return. A larger $r_{\text{Omega}}$ indicates stronger dominance of gains over losses, thus providing a more comprehensive characterization of portfolio quality.

\subsection{Results and Discussion}\label{Results}

\begin{table}[htbp]
\centering
\caption{Backtest 1, performance comparison across strategies (best values in each column are highlighted in bold).}
\label{tab:Backtest1}
\small
\begin{tabular}{lcccc}
\toprule
\textbf{Strategy} & \textbf{Return} & \textbf{Sharpe Ratio} & \textbf{Sortino Ratio} & \textbf{Omega Ratio} \\
\midrule
SAMP-HDRL & 0.5652 & 0.1720 & 0.1580 & 1.6025 \\
CRP       & 0.3251 & 0.1277 & 0.1043 & 1.4194 \\
UBAH      & 0.3226 & 0.1274 & 0.1050 & 1.4180 \\
M0        & 0.3152 & 0.1122 & 0.0926 & 1.3600 \\
UP        & 0.3250 & 0.1278 & 0.1044 & 1.4196 \\
EG        & 0.3175 & 0.1233 & 0.1031 & 1.4005 \\
PAMR      & -0.5208 & -0.0930 & -0.0834 & 0.7708 \\
CWMR      & -0.5307 & -0.0924 & -0.0828 & 0.7735 \\
CORNK     & -0.1476 & -0.0407 & -0.0452 & 0.8960 \\
CAPM      & -0.5079 & -0.1618 & -0.1620 & 0.6447 \\
EIIE      & 0.3112 & 0.1213 & 0.1010 & 1.4026 \\
FinRL     & 0.1380 & 0.0215 & 0.0156 & 1.0813 \\
EST       & 0.2384 & 0.0987 & 0.0881 & 1.3334 \\
II        & 0.0878 & 0.0175 & 0.0102 & 1.0499 \\
SARL      & 0.2299 & 0.0911 & 0.0712 & 1.2841 \\
PPO       & 0.2893 & 0.1103 & 0.0907 & 1.3619 \\
PNN       & 0.0062 & 0.0014 & -0.0059 & 1.0040 \\
DPG       & 0.1617 & 0.0382 & 0.0300 & 1.1057 \\
TARN      & 0.3195 & 0.1256 & 0.1032 & 1.4118 \\
DeepMPT   & 0.2349 & 0.0519 & 0.0422 & 1.1680 \\
LSRE-CAAN & 0.3169 & 0.1242 & 0.1033 & 1.4041 \\
FTRL      & \textbf{0.6722} & \textbf{0.1829} & \textbf{0.2379} & \textbf{1.6523} \\
\bottomrule
\end{tabular}
\end{table}

\begin{figure}[htbp]
    \centering
    \includegraphics[width=8cm,height=4cm]{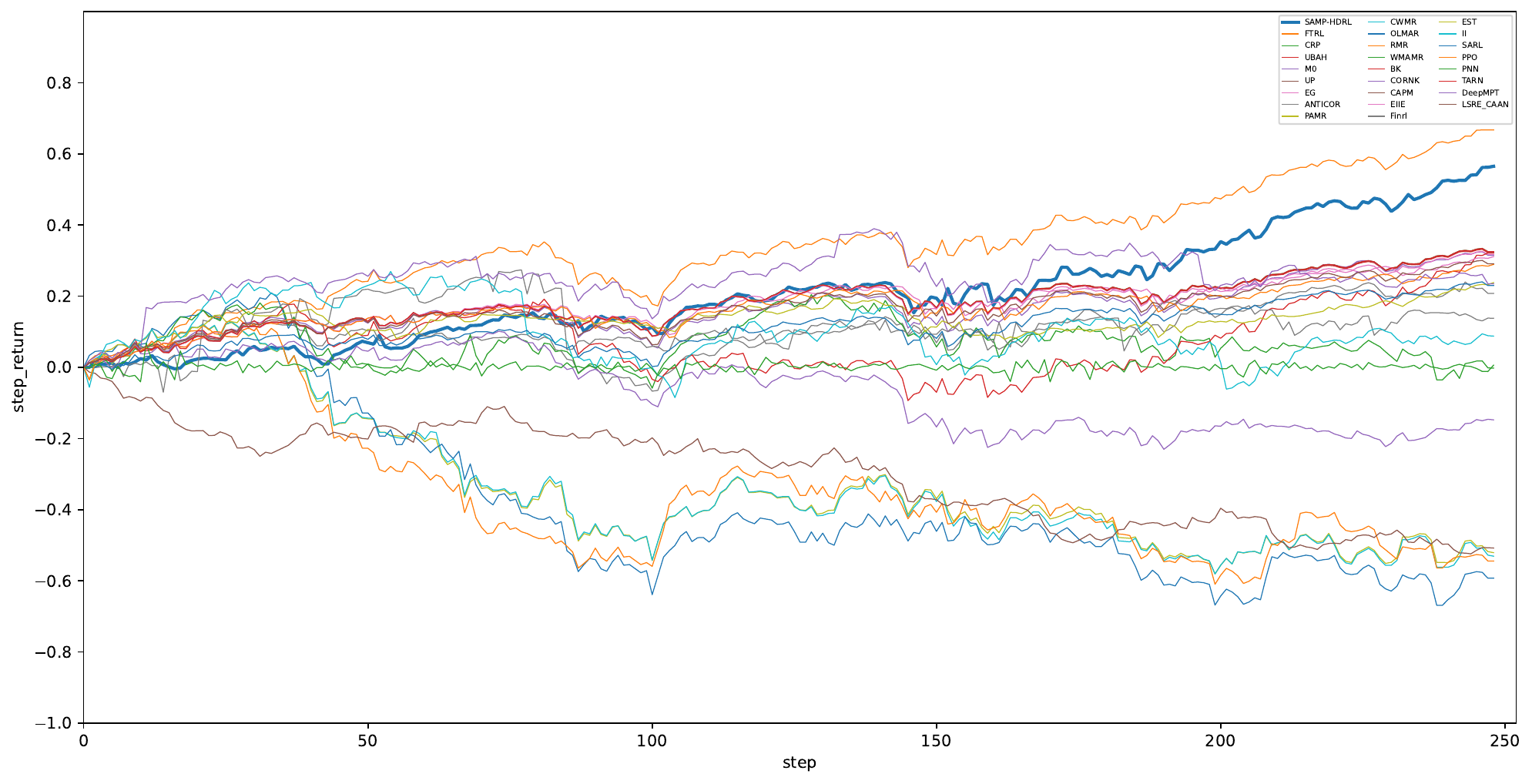}
    \caption[Daily cumulative return rate for Backtest 1.]{
    \textbf{Daily cumulative return rate for Backtest 1.} It depicts the volatility of FTRL cumulative returns and the comparative strategy in Backtest 1.}
    \label{Backtest1}
\end{figure}

\begin{table}[htbp]
\centering
\caption{Backtest 2, performance comparison across strategies (best values in each column are highlighted in bold).}
\label{tab:Backtest2}
\small
\begin{tabular}{lcccc}
\toprule
\textbf{Strategy} & \textbf{Return} & \textbf{Sharpe Ratio} & \textbf{Sortino Ratio} & \textbf{Omega Ratio} \\
\midrule
SAMP-HDRL & \textbf{0.7047} & \textbf{0.0816} & \textbf{0.0712} & \textbf{1.2744} \\
CRP       & 0.1591 & 0.0196 & 0.0145 & 1.0659 \\
UBAH      & 0.0990 & 0.0121 & 0.0076 & 1.0402 \\
M0        & 0.1669 & 0.0202 & 0.0154 & 1.0667 \\
UP        & 0.1580 & 0.0195 & 0.0144 & 1.0655 \\
EG        & 0.0936 & 0.0121 & 0.0073 & 1.0402 \\
PAMR      & -1.3773 & -0.0996 & -0.0882 & 0.7020 \\
CWMR      & -1.3641 & -0.0986 & -0.0874 & 0.7047 \\
CORNK     & -0.6349 & -0.0595 & -0.0633 & 0.8413 \\
CAPM      & -0.7527 & -0.1299 & -0.1181 & 0.6504 \\
EIIE      & 0.1261 & 0.0165 & 0.0116 & 1.0553 \\
FinRL     & 0.0817 & 0.0048 & 0.0029 & 1.0178 \\
EST       & 0.0111 & 0.0026 & -0.0040 & 1.0075 \\
II        & 0.4620 & 0.0460 & 0.0431 & 1.1424 \\
SARL      & 0.0336 & 0.0040 & 0.0005 & 1.0128 \\
PPO       & 0.0994 & 0.0126 & 0.0081 & 1.0419 \\
PNN       & -0.0094 & -0.0005 & -0.0022 & 0.9983 \\
DPG       & -0.2163 & -0.0152 & -0.0169 & 0.9528 \\
TARN      & 0.1553 & 0.0190 & 0.0140 & 1.0640 \\
DeepMPT   & -0.0728 & -0.0059 & -0.0078 & 0.9819 \\
LSRE-CAAN & 0.1897 & 0.0231 & 0.0177 & 1.0780 \\
FTRL      & 0.5292 & 0.0606 & 0.0519 & 1.2059 \\
\bottomrule
\end{tabular}
\end{table}

\begin{figure}[htbp]
    \centering
    \includegraphics[width=8cm,height=4cm]{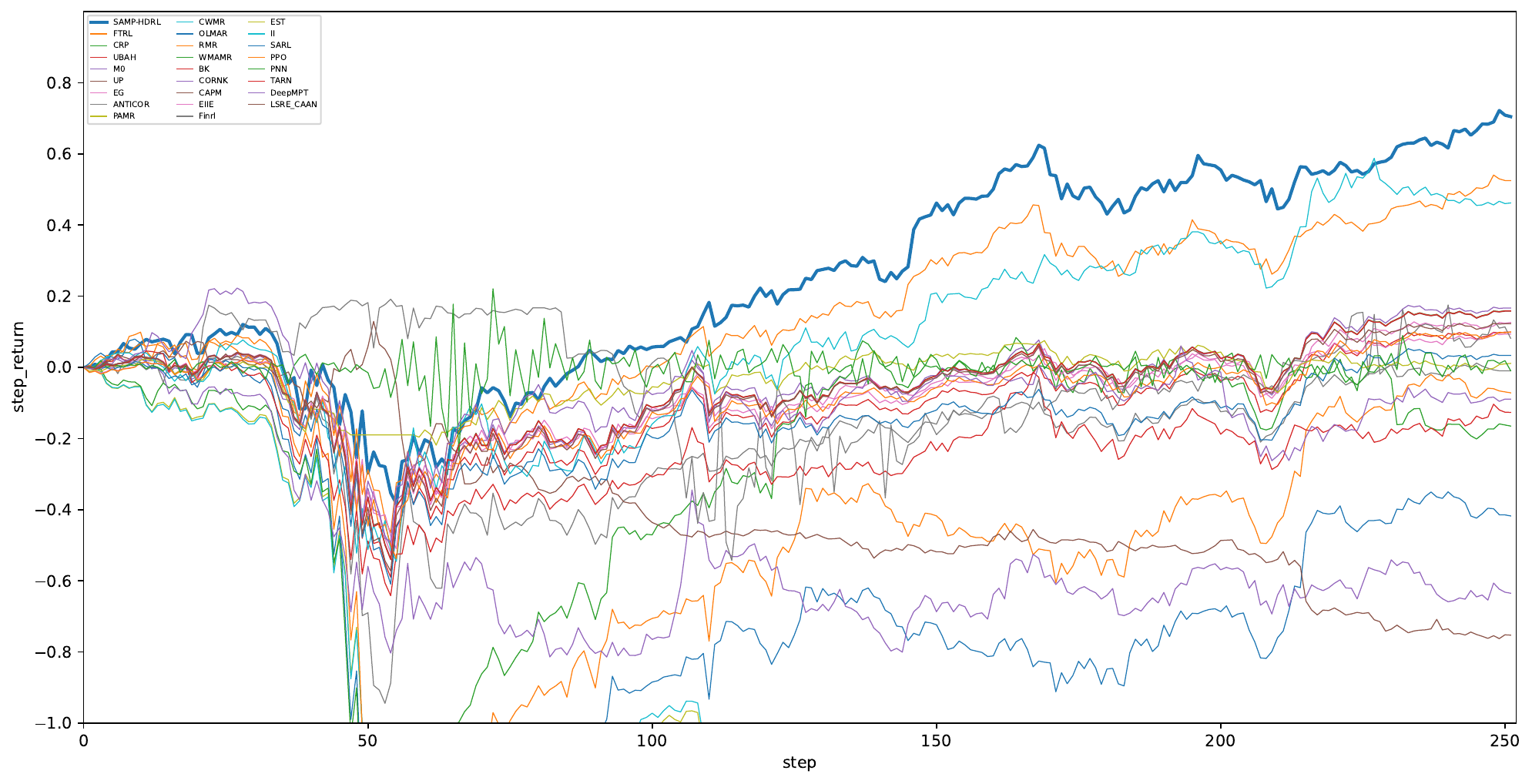}
    \caption[Daily cumulative return rate for Backtest 2.]{
    \textbf{Daily cumulative return rate for Backtest 2.} It depicts the volatility of FTRL cumulative returns and the comparative strategy in Backtest 2.}
    \label{Backtest2}
\end{figure}

\begin{table}[htbp]
\centering
\caption{Backtest 3, performance comparison across strategies (best values in each column are highlighted in bold).}
\label{tab:Backtest3}
\small
\begin{tabular}{lcccc}
\toprule
\textbf{Strategy} & \textbf{Return} & \textbf{Sharpe Ratio} & \textbf{Sortino Ratio} & \textbf{Omega Ratio} \\
\midrule
SAMP-HDRL & \textbf{0.3938} & \textbf{0.1101} & \textbf{0.1004} & \textbf{1.3328} \\
CRP       & 0.2795 & 0.1032 & 0.0899 & 1.3129 \\
UBAH      & 0.2798 & 0.1023 & 0.0889 & 1.3092 \\
M0        & 0.2255 & 0.0721 & 0.0606 & 1.2059 \\
UP        & 0.2788 & 0.1030 & 0.0899 & 1.3121 \\
EG        & 0.2612 & 0.0931 & 0.0788 & 1.2799 \\
PAMR      & -0.7418 & -0.1372 & -0.1340 & 0.6862 \\
CWMR      & -0.7830 & -0.1403 & -0.1383 & 0.6793 \\
CORNK     & -0.4646 & -0.0909 & -0.0868 & 0.7682 \\
CAPM      & -0.4190 & -0.1209 & -0.1215 & 0.6997 \\
EIIE      & 0.2269 & 0.0880 & 0.0736 & 1.2625 \\
FinRL     & 0.2018 & 0.0252 & 0.0177 & 1.1093 \\
EST       & 0.1626 & 0.0525 & 0.0416 & 1.1463 \\
II        & 0.1828 & 0.0455 & 0.0371 & 1.1286 \\
SARL      & 0.1607 & 0.0520 & 0.0408 & 1.1425 \\
PPO       & 0.2253 & 0.0860 & 0.0729 & 1.2543 \\
PNN       & 0.0034 & 0.0008 & -0.0064 & 1.0020 \\
DPG       & 0.1337 & 0.0253 & 0.0218 & 1.0700 \\
TARN      & 0.2798 & 0.1029 & 0.0894 & 1.3130 \\
DeepMPT   & 0.0451 & 0.0079 & 0.0024 & 1.0248 \\
LSRE-CAAN & 0.2813 & 0.1025 & 0.0892 & 1.3121 \\
FTRL      & 0.3746 & 0.1048 & 0.0949 & 1.3070 \\
\bottomrule
\end{tabular}
\end{table}

\begin{figure}[htbp]
    \centering
    \includegraphics[width=8cm,height=4cm]{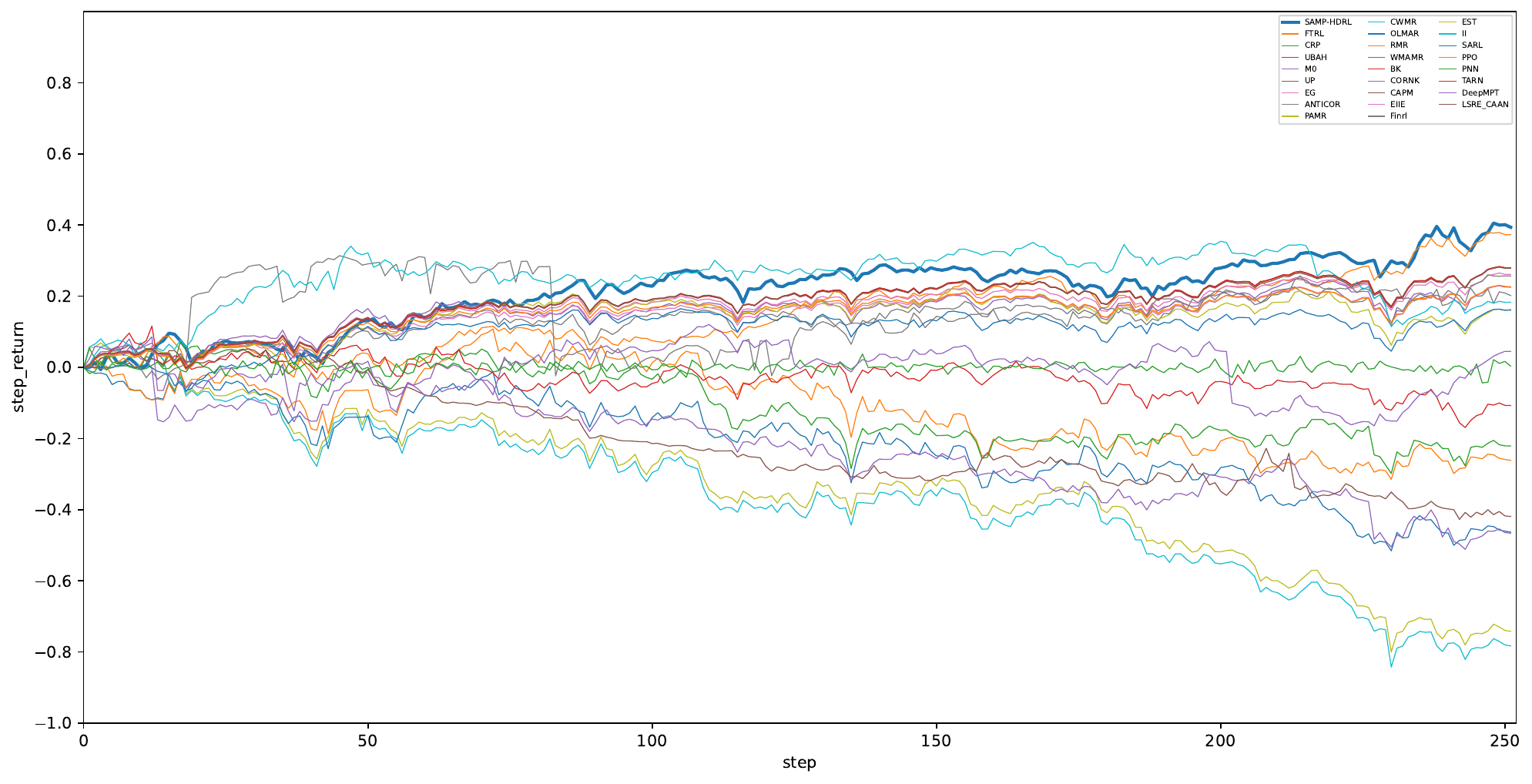}
    \caption[Daily cumulative return rate for Backtest 3.]{
    \textbf{Daily cumulative return rate for Backtest 3.} It depicts the volatility of FTRL cumulative returns and the comparative strategy in Backtest 3.}
    \label{Backtest3}
\end{figure}

We obtained the results in Tables~\cref*{tab:Backtest1,tab:Backtest2,tab:Backtest3} by conducting experiments on data sets. Daily cumulative returns of strategies are given in ~\Cref*{Backtest1,Backtest2,Backtest3}. In Backtest 1, SAMP-HDRL outperforms all strategies except FTRL, achieving at least 74\%, 34\%, 52\%, and 13\% higher values in Return, Sharpe Ratio, Sortino Ratio, and Omega Ratio, respectively. In Backtest 2, SAMP-HDRL demonstrates clear superiority over all strategies, with improvements of at least 33\%, 34\%, 36\%, and 5\% across the four metrics. In Backtest 3, SAMP-HDRL still maintains an advantage, exceeding all other methods by at least 5\%, 5\%, 5\%, and 2\% in Return, Sharpe Ratio, Sortino Ratio, and Omega Ratio, respectively.  

It is noteworthy that the time period corresponding to Backtest 1 represents a relatively stable and upward-trending market environment \cite{bib88,bib89}. Within this setting, SAMP-HDRL underperforms FTRL by approximately $16\%$, highlighting the sensitivity of the proposed framework to persistently bullish market regimes. This observation can be attributed to two key factors. First, FTRL does not incorporate explicit risk management, thereby allowing the strategy to fully exploit upward trends and maximize profitability in such conditions \cite{bib15}. Second, the hierarchical architecture of SAMP-HDRL not only captures temporal patterns but also integrates structural mechanisms for risk balancing and robustness \cite{bib48}. This design creates a dynamic trade-off between return maximization and risk control. Under relatively calm and upward-trending conditions, this trade-off may suppress certain high-risk, high-reward opportunities, resulting in marginally lower returns compared to the purely return-oriented FTRL \cite{bib90,bib91}. Nevertheless, when market volatility intensifies or uncertainty becomes dominant \cite{bib58}, this mechanism is more advantageous, as evidenced by the superior performance of SAMP-HDRL in Backtest 2. 
\begin{table}[htbp]
\centering
\caption[Comparison of strategies]{Comparison of FTRL, FTRL-risk, and SAMP-HDRL across three backtest datasets (best values in each column are highlighted in bold).}
\label{tab:ftrl-risk}
\scriptsize 
\renewcommand{\arraystretch}{1.1} 
\setlength{\tabcolsep}{5pt} 
\begin{tabular}{llcccc}
\toprule
\textbf{Backtest} & \textbf{Strategy} & \textbf{Return} & \textbf{Sharpe Ratio} & \textbf{Sortino Ratio} & \textbf{Omega Ratio} \\
\midrule
\multirow{3}{*}{Backtest 1} 
& FTRL-risk  & 0.5167 & 0.1572 & 0.1454 & 1.5402 \\
& FTRL       & \textbf{0.6722} & \textbf{0.1829} & \textbf{0.2379} & \textbf{1.6523} \\
& SAMP-HDRL  & 0.5652 & 0.1720 & 0.1580 & 1.6025 \\
\midrule
\multirow{3}{*}{Backtest 2} 
& FTRL-risk  & 0.6866 & 0.0794 & 0.0710 & 1.2672 \\
& FTRL       & 0.5292 & 0.0606 & 0.0519 & 1.2059 \\
& SAMP-HDRL  & \textbf{0.7047} & \textbf{0.0816} & \textbf{0.0712} & \textbf{1.2744} \\
\midrule
\multirow{3}{*}{Backtest 3} 
& FTRL-risk  & 0.3916 & 0.1095 & 0.0970 & 1.3316 \\
& FTRL       & 0.3746 & 0.1048 & 0.0949 & 1.3070 \\
& SAMP-HDRL  & \textbf{0.3938} & \textbf{0.1101} & \textbf{0.1004} & \textbf{1.3328} \\
\bottomrule
\end{tabular}
\end{table}

To further validate the findings, an additional set of experiments is conducted with a variant of FTRL, denoted as FTRL-risk. This extension augments the original FTRL by incorporating a lower-level module for dynamic asset classification and capital allocation, consistent with the design of SAMP-HDRL. The three backtest datasets are used, and the results are summarized in Table~\ref{tab:ftrl-risk}, which reports the performance of FTRL, its risk-augmented variant, and SAMP-HDRL. In the time period corresponding to Backtest 1, characterized by a relatively stable and upward-trending market, FTRL achieves the best outcomes across all metrics, surpassing SAMP-HDRL by $19\%$, $6\%$, $51\%$, and $3\%$ in Return, Sharpe ratio, Sortino ratio, and Omega ratio, respectively. By contrast, FTRL-risk underperforms FTRL, with the same metrics being lower by $23\%$, $14\%$, $39\%$, and $7\%$, respectively, confirming the inference that explicit risk control may suppress high-risk, high-reward opportunities in bullish regimes \cite{bib90,bib91}. In the period corresponding to Backtest 2, marked by heightened volatility \cite{bib58}, SAMP-HDRL consistently outperforms all strategies, exceeding FTRL by $33\%$, $35\%$, $37\%$, and $6\%$ in the four metrics, respectively. FTRL-risk also improves upon FTRL with gains of $30\%$, $31\%$, $37\%$, and $5\%$, respectively, indicating that risk-sensitive modifications enhance profitability and risk-adjusted performance, though still falling short of the robustness achieved by SAMP-HDRL. In the time horizon of Backtest 3, reflecting sideways and mixed conditions \cite{bib93}, SAMP-HDRL again maintains the lead, outperforming FTRL by $5\%$, $5\%$, $6\%$, and $2\%$ in Return, Sharpe ratio, Sortino ratio, and Omega ratio, respectively, while FTRL-risk shows modest improvements of $5\%$, $4\%$, $2\%$, and $2\%$ over FTRL, respectively, yet remains inferior to SAMP-HDRL overall. Overall, these results reinforce the prior analysis: while FTRL benefits most in stable upward markets due to the absence of explicit risk constraints, the integration of risk control substantially strengthens resilience in volatile and uncertain environments, enabling SAMP-HDRL to achieve a more effective dynamic balance between return maximization and risk management.

\begin{table}[htbp]
\centering
\caption[Ablation study]{Ablation study of SAMP-HDRL, performance comparison with different modules removed across three backtest datasets (best values in each column are highlighted in bold).}
\label{tab:ablation_full}
\footnotesize
\renewcommand{\arraystretch}{1.1}
\setlength{\tabcolsep}{5pt}
\begin{tabular}{llcccc}
\toprule
\textbf{Backtest} & \textbf{Strategy} & \textbf{Return} & \textbf{Sharpe Ratio} & \textbf{Sortino Ratio} & \textbf{Omega Ratio} \\
\midrule
\multirow{5}{*}{Backtest 1} 
& SAMP-HDRL w/o upper & 0.4391 & 0.1479 & 0.1306 & 1.4917 \\
& SAMP-HDRL w/o lower & \textbf{0.6722} & \textbf{0.1829} & \textbf{0.2379} & \textbf{1.6523} \\
& SAMP-HDRL w/o dc    & 0.2639 & 0.1025 & 0.0820 & 1.3135 \\
& SAMP-HDRL w/o ca    & 0.4545 & 0.1570 & 0.1370 & 1.5408 \\
& SAMP-HDRL           & 0.5652 & 0.1720 & 0.1580 & 1.6025 \\
\midrule
\multirow{5}{*}{Backtest 2} 
& SAMP-HDRL w/o upper & 0.5103 & 0.0611 & 0.0512 & 1.2110 \\
& SAMP-HDRL w/o lower & 0.5292 & 0.0606 & 0.0519 & 1.2059 \\
& SAMP-HDRL w/o dc    & 0.6284 & 0.0713 & 0.0647 & 1.2363 \\
& SAMP-HDRL w/o ca    & 0.4679 & 0.0553 & 0.0467 & 1.1897 \\
& SAMP-HDRL           & \textbf{0.7047} & \textbf{0.0816} & \textbf{0.0712} & \textbf{1.2744} \\
\midrule
\multirow{5}{*}{Backtest 3} 
& SAMP-HDRL w/o upper & 0.3265 & 0.0977 & 0.0897 & 1.2887 \\
& SAMP-HDRL w/o lower & 0.3746 & 0.1048 & 0.0949 & 1.3070 \\
& SAMP-HDRL w/o dc    & 0.3895 & 0.1118 & 0.1047 & \textbf{1.3354} \\
& SAMP-HDRL w/o ca    & 0.3481 & 0.1099 & 0.1017 & 1.3352 \\
& SAMP-HDRL           & \textbf{0.3938} & \textbf{0.1101} & \textbf{0.1004} & 1.3328 \\
\bottomrule
\end{tabular}
\end{table}

To further validate the effectiveness and necessity of the proposed innovations in SAMP-HDRL---namely upper–lower coordination, dynamic clustering, and capital allocation---we design a set of ablation experiments. Specifically, four variants of SAMP-HDRL are constructed: SAMP-HDRL w/o upper, SAMP-HDRL w/o lower, SAMP-HDRL w/o dc, and SAMP-HDRL w/o ca, which remove the upper-level framework, the lower-level agents, the dynamic clustering module, and the capital allocation module, respectively. To ensure operability, in SAMP-HDRL w/o upper the upper framework is replaced with the UBAH \cite{bib73} strategy to provide global decision outputs, thereby testing its contribution to inter-asset correlation modeling. In SAMP-HDRL w/o lower, the entire lower-level agent structure is removed to examine its role in intra-asset temporal modeling and risk control. In SAMP-HDRL w/o dc, the dynamic clustering mechanism is replaced by a static clustering procedure performed before training, verifying its advantage in adapting to evolving market structures. Finally, in SAMP-HDRL w/o ca, the original capital allocation mechanism is replaced by equal allocation between asset groups, in order to assess its contribution to optimizing the risk–return trade-off. The empirical results are reported in Table~\ref{tab:ablation_full}.  

Table~\ref{tab:ablation_full} reports the performance of SAMP-HDRL and its ablated variants across the three backtest datasets. The results demonstrate that the proposed innovations---upper–lower coordination, dynamic clustering, and capital allocation---play indispensable roles in enhancing overall performance.  

In the time period corresponding to Backtest 1, SAMP-HDRL outperforms the variant without the upper framework by $29\%$, $16\%$, $21\%$, and $7\%$ in Return, Sharpe ratio, Sortino ratio, and Omega ratio, respectively. Relative to the variant without dynamic clustering, the improvements are $114\%$, $68\%$, $93\%$, and $22\%$, respectively, while relative to the variant without capital allocation, the gains are $24\%$, $10\%$, $15\%$, and $4\%$, respectively. However, the variant without the lower agents achieves a Return higher than SAMP-HDRL by $19\%$, confirming the earlier inference that risk-control mechanisms may restrict profit maximization under one-sided upward conditions \cite{bib90,bib91}.  

In the interval of Backtest 2, characterized by heightened volatility, SAMP-HDRL demonstrates consistent superiority. Compared with the variant without the upper framework, the improvements are $38\%$, $34\%$, $39\%$, and $5\%$ in the four metrics, respectively. Relative to the variant without the lower framework, the gains are $33\%$, $35\%$, $37\%$, and $6\%$, respectively. Against the variant without dynamic clustering, the improvements are $12\%$, $14\%$, $10\%$, and $3\%$, respectively, while relative to the variant without capital allocation, the margins rise to $51\%$, $47\%$, $53\%$, and $7\%$, respectively. These results underscore that all three mechanisms contribute indispensably to robustness and the risk–return balance under turbulent conditions.  

In the horizon of Backtest 3, reflecting sideways and mixed market conditions, SAMP-HDRL outperforms the variant without the upper framework by $21\%$, $13\%$, $12\%$, and $3\%$ in Return, Sharpe ratio, Sortino ratio, and Omega ratio, respectively, and exceeds the variant without the lower framework by $5\%$, $1\%$, $2\%$, and $2\%$, respectively. Relative to the variant without capital allocation, the improvements are $13\%$, $0.1\%$, $-1\%$, and $-0.2\%$ in the same metrics, respectively. When compared with the variant without dynamic clustering, SAMP-HDRL achieves an increase of $1\%$ in Return but decreases of $2\%$ and $0.3\%$ in Sharpe and Sortino ratios, respectively, while its Omega ratio is slightly lower by $0.2\%$. This indicates that dynamic clustering is particularly effective in optimizing the risk–return trade-off in non-trending markets.  

Overall, the ablation study validates the effectiveness of the three key innovations: the upper framework substantially strengthens robustness through inter-asset correlation modeling, the lower agents are indispensable for risk control in volatile markets, and dynamic clustering together with capital allocation effectively optimize the risk–return balance in structurally shifting and oscillating environments, thereby underpinning the overall superiority of SAMP-HDRL.  

SHapley Additive exPlanations (SHAP) \cite{bib92} offers a cooperative game–theoretic framework for interpretability by attributing the marginal contribution of each input feature to the model output in a fair manner. In this study, SHAP is integrated into the DRL-based portfolio allocation framework to quantify the influence of each asset’s historical price trajectory on the assigned portfolio weights. Specifically, for each backtest period, SHAP values are computed for the actions of both lower-level agents, capturing how past market movements and clustering masks contribute to allocation decisions.

\begin{figure}[htbp]
    \centering
    \includegraphics[width=12cm]{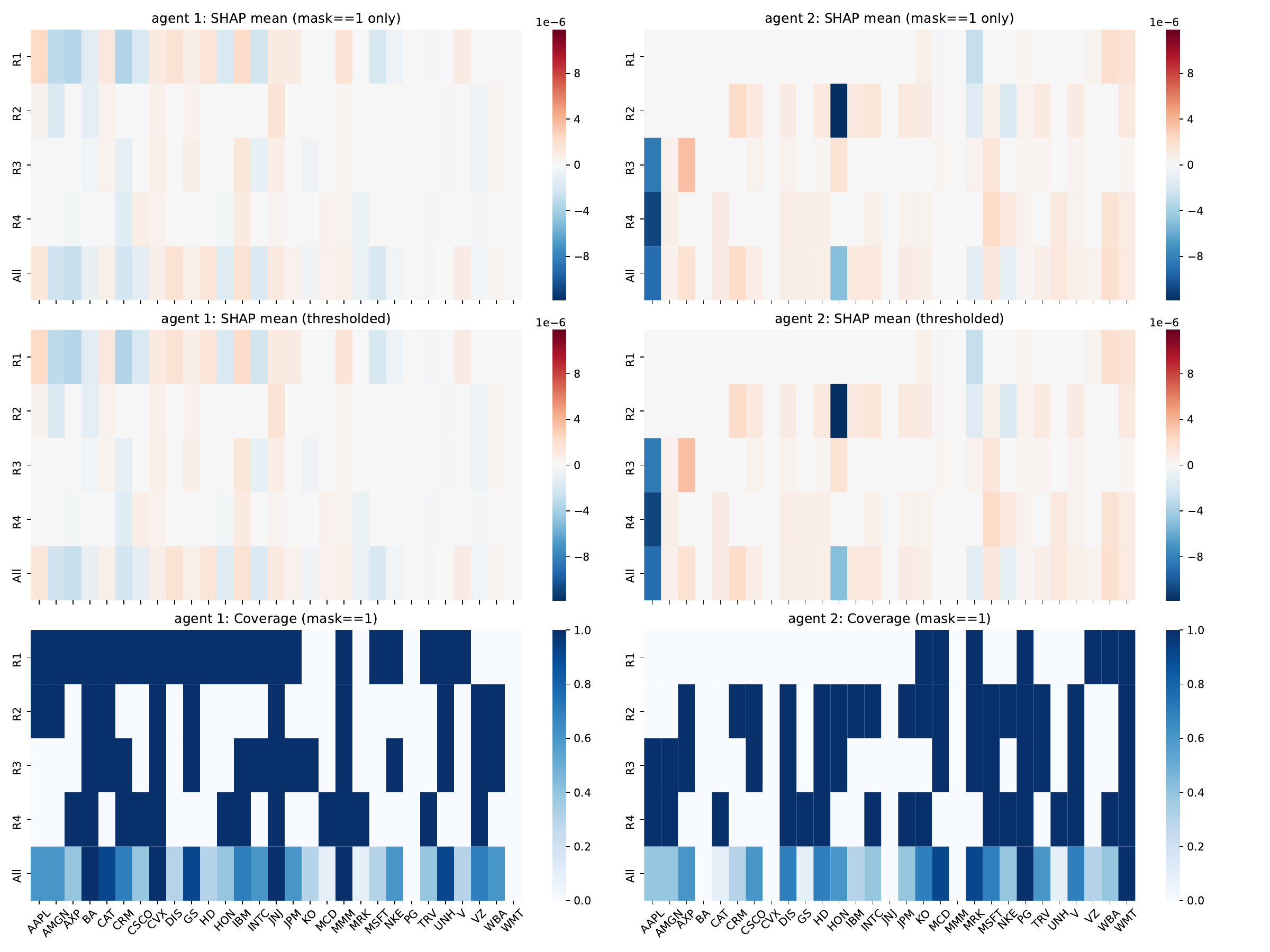}
    \caption[SHAP analysis 1.]{
    \textbf{SHAP analysis 1.}SHAP analysis for Backtest 1 (2019), showing mean, thresholded, and coverage visualizations. Agent 1 (remaining assets) provides broad diversification, while Agent 2 (high-quality assets) focuses on core contributors.}
    \label{SHAP1}
\end{figure}
\begin{figure}[htbp]
    \centering
    \includegraphics[width=12cm]{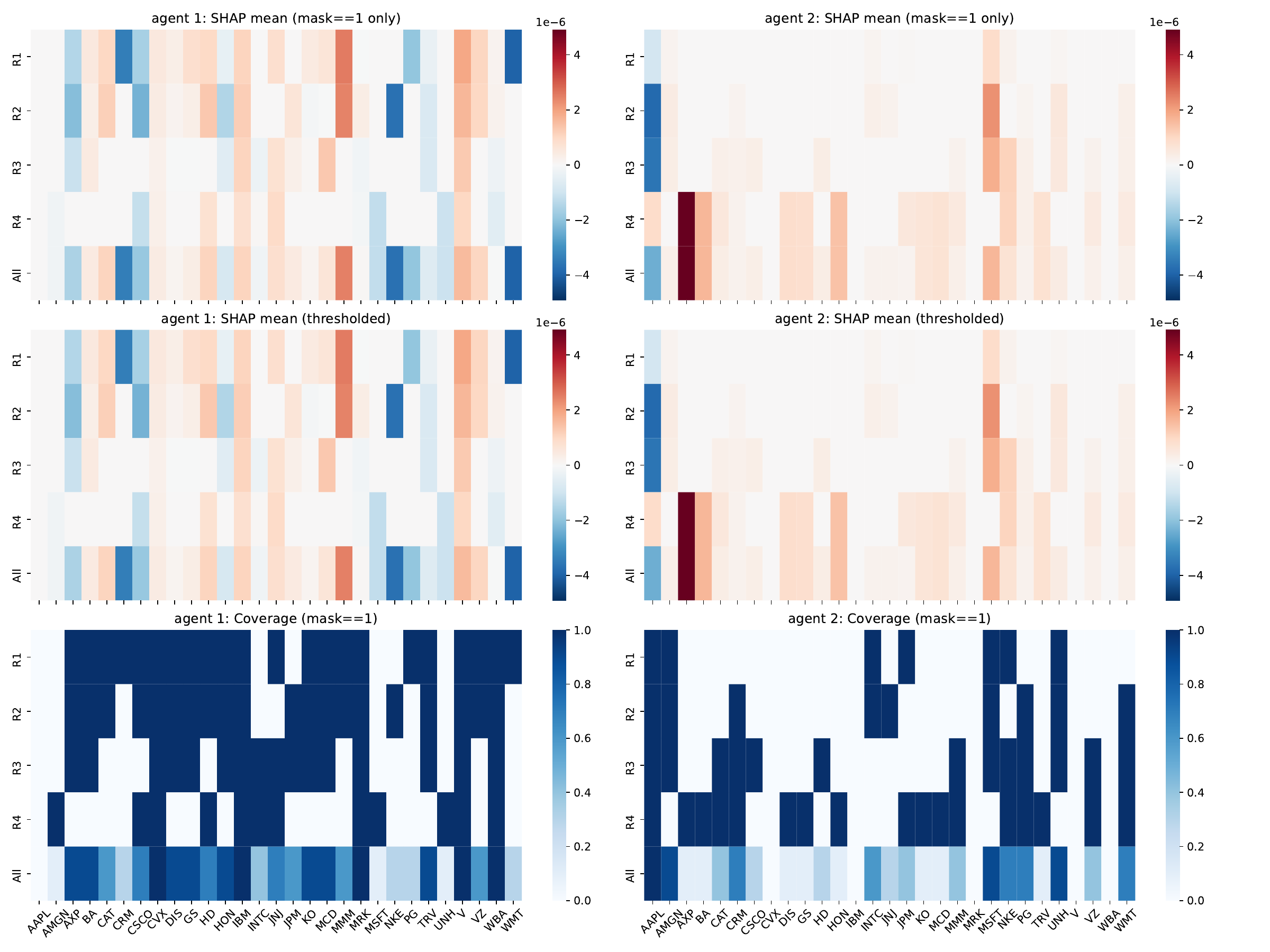}
    \caption[SHAP analysis 2.]{
    \textbf{SHAP analysis 2.}SHAP analysis for Backtest 2 (2020), showing mean, thresholded, and coverage visualizations. Agent 1 (remaining assets) provides broad diversification, while Agent 2 (high-quality assets) focuses on core contributors.}
    \label{SHAP2}
\end{figure}
\begin{figure}[htbp]
    \centering
    \includegraphics[width=12cm]{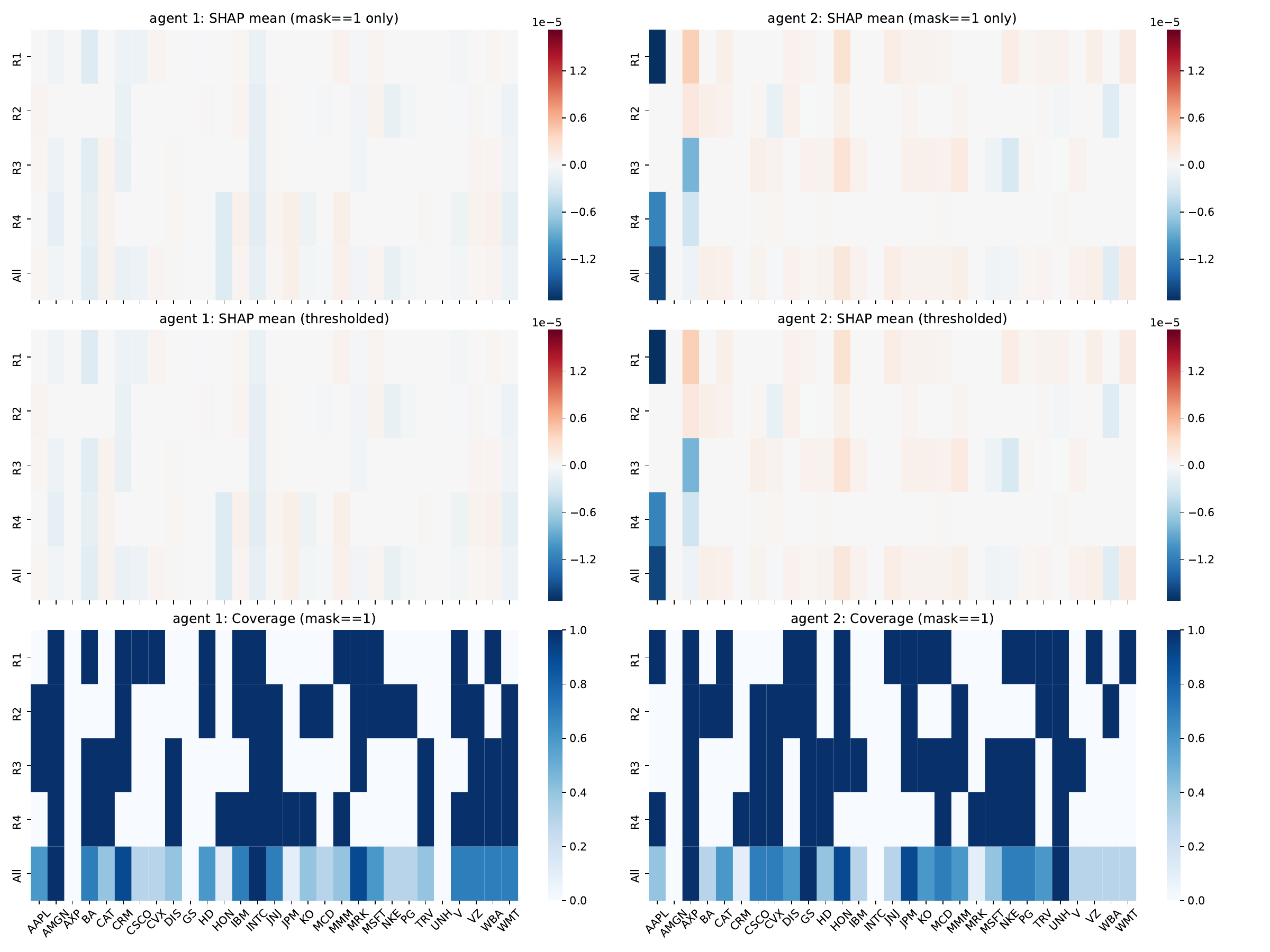}
    \caption[SHAP analysis 3.]{
    \textbf{SHAP analysis 3.}SHAP analysis for Backtest 3 (2021), showing mean, thresholded, and coverage visualizations. Agent 1 (remaining assets) provides broad diversification, while Agent 2 (high-quality assets) focuses on core contributors.}
    \label{SHAP3}
\end{figure}

The resulting visualizations (\Cref*{SHAP1,SHAP2,SHAP3}) consist of three subplots corresponding to distinct market regimes. In each subplot, the vertical axis represents the temporal dimension (trading days), while the horizontal axis corresponds to individual assets. The panels labeled R1–R4 correspond to the four post-reclassification periods in which cluster memberships remain stable, and ALL provides the aggregated view across the full year. The color intensity encodes the SHAP value magnitude, indicating each asset’s contribution to portfolio weight decisions at a given time. Warm colors denote positive contributions that increase allocation weights, whereas cool colors indicate negative contributions leading to reduced exposure. The superimposed trend lines highlight periods of regime transition or allocation shifts, allowing direct visual comparison of agent attention across market phases.

To capture the model’s explanatory behavior from multiple perspectives, three complementary SHAP visualizations are presented. The mean SHAP plot reflects the overall magnitude of asset contributions by averaging absolute SHAP values over time, thus revealing globally dominant drivers of portfolio allocation. The thresholded SHAP visualization focuses on periods where SHAP magnitudes exceed a specified significance level, filtering out minor fluctuations and emphasizing decisive yet infrequent allocation impulses. In contrast, the coverage SHAP plot quantifies how frequently each asset exhibits substantial influence, thereby capturing the persistence and stability of its contribution across trading periods. Together, these three visual perspectives jointly describe both the intensity and temporal prevalence of feature impacts, enabling a more comprehensive interpretation of agent decision dynamics.

Examination of these SHAP patterns further reveals that the two lower-level agents exhibit complementary behaviors: Agent~1 maintains wide coverage across assets to ensure diversification, while Agent~2 concentrates on high-quality clusters that dominate returns during key uptrends. The temporal synchronization between these two attention patterns provides empirical evidence that the hierarchical coordination mechanism effectively adapts to varying market regimes, thereby enhancing both interpretability and transparency of the overall DRL decision process.

~\Cref*{SHAP1,SHAP2,SHAP3} presents the SHAP interpretability results of SAMP-HDRL across Backtest 1 (2019), Backtest 2 (2020), and Backtest 3 (2021), corresponding to a stable upward market, a non-stationary market, and a recovery phase with oscillating dynamics, respectively. Overall, the two lower-level agents exhibit a clear division of roles: Agent~1, which manages the remaining assets, maintains broad coverage to ensure diversification, while Agent~2, which focuses on high-quality assets, allocates concentrated attention to the core drivers of portfolio performance. This “diversified + concentrated” coordination adapts dynamically to varying market regimes, thereby enhancing the interpretability of an otherwise black-box DRL policy.  

In the stable upward market of 2019 \cite{bib88,bib89}, Agent~1 distributes attention broadly across assets to sustain diversification, whereas Agent~2 concentrates on high-quality sectors such as technology and finance. Both the SHAP mean and thresholded analyses highlight the persistent positive contributions of these core assets during key upward phases. This complementary allocation aligns with the market’s unidirectional growth, enabling the model to capture structural gains while preserving robustness.  

In the non-stationary market of 2020 \cite{bib58}, the COVID-19 shock induces abrupt shifts in asset risk exposures. The SHAP mean results reveal frequent switches between positive and negative contributions, reflecting adaptive responses to heightened uncertainty. Agent~1 continues to maintain broad coverage to buffer systemic risk, while Agent~2 sharply narrows its focus to a small set of high-confidence quality assets, capturing rebounds and structural opportunities. The thresholded SHAP results confirm that the model effectively filters noise in this turbulent environment, retaining only decisive features and thereby demonstrating robustness under non-stationarity.  

In the oscillating recovery of 2021 \cite{bib93}, the market exhibits upward tendencies accompanied by heightened volatility and uncertainty. The SHAP mean results show relatively smaller marginal contributions across assets, indicating weaker single-stock drivers. Agent~1 maintains balanced coverage across most assets to mitigate risks, whereas Agent~2 selectively emphasizes a few quality assets during specific intervals, forming concentrated allocations. The thresholded SHAP analysis further validates a “broad coverage + selective focus” strategy, balancing risk and return in a sideways environment.  

Taken together, the SHAP analyses across three regimes demonstrate that the two lower-level agents do not allocate weights arbitrarily but instead operate through a complementary mechanism. Agent~1 ensures diversification across remaining assets, while Agent~2 concentrates on high-quality assets. This coordination enables the policy to emphasize structural drivers in upward markets, reinforce risk buffering in non-stationary regimes, and integrate diversification with selective focus in oscillating recoveries. Such behavior not only substantiates the rationality of the agents’ decisions but also enhances the interpretability and credibility \cite{bib118} of DRL-based portfolio optimization.

\section{Conclusions and Future Plan}\label{Conclusions}

This study introduces SAMP-HDRL, a novel hierarchical deep reinforcement learning framework for portfolio management that integrates upper–lower agent coordination, dynamic clustering, and capital allocation. Extensive experiments on three backtest datasets demonstrate that SAMP-HDRL consistently and significantly outperforms both traditional baselines and recent DRL methods across return, Sharpe ratio, Sortino ratio, and Omega ratio. To the best of our knowledge, we are the first to provide a comprehensive integration of hierarchical coordination, clustering, and capital allocation within a unified end-to-end paradigm for portfolio optimization. Comparative and ablation analyses confirm the indispensable contributions of each module: the upper framework enhances inter-asset correlation modeling, the lower agents strengthen risk control, and the combination of clustering with capital allocation ensures an effective balance between diversification and focused allocation. Moreover, SHAP-based interpretability analysis reveals that the two lower-level agents operate in a complementary manner, with Agent~1 supporting diversification across remaining assets and Agent~2 concentrating on high-quality assets, thereby offering transparent and economically consistent explanations of portfolio decisions.

Nevertheless, several limitations warrant careful consideration. First, the hierarchical architecture is not strictly end-to-end, as the upper- and lower-level agents are optimized in a staged manner, which may constrain full cross-level information integration. Second, although dynamic clustering enhances intra-group feature representation, the framework does not explicitly capture inter-cluster dependencies, thereby risking the omission of valuable cross-asset correlations. Third, the interpretability analysis primarily relies on SHAP, which provides post-hoc rather than real-time or causality-aware explanations, leaving open questions regarding transparent monitoring of the decision process in live trading environments. Finally, the evaluation is restricted to price-based time series from the DJIA, without the inclusion of macroeconomic, sentiment, or alternative cross-market signals, which may limit generalizability to broader financial contexts. Collectively, these issues underscore both methodological and practical dimensions in which the framework can be further advanced.  

Future research will seek to address these challenges along several directions. First, more integrated end-to-end optimization paradigms will be explored, enabling upper- and lower-level agents to co-adapt through shared objectives and thereby strengthen cross-level synergy. Second, graph neural networks and correlation-aware attention mechanisms will be incorporated to explicitly model inter-cluster dependencies and improve diversification. Third, interpretability will be extended toward real-time and causality-aware frameworks, facilitating continuous monitoring of portfolio decisions and validation under counterfactual scenarios. Fourth, the framework will be enriched with multi-modal and cross-market features---including macroeconomic indicators, textual sentiment, and alternative signals---thereby improving robustness under distributional shifts. Finally, future extensions will emphasize scalability to larger universes and higher-frequency trading environments, coupled with empirical validation under realistic constraints such as transaction costs, liquidity limitations, and execution slippage. These research avenues aim to further advance the practicality, adaptability, and theoretical depth of hierarchical deep reinforcement learning for portfolio management.

\section{Declarations}\label{Declarations}

The authors did not receive funding from any organization for the submitted work. The authors have no relevant financial or non-financial interests to disclose. 

Xiaotian Ren determined the experimental design, wrote code, participated in conducting experiments, analyzed data, created graphics, and drafted the manuscript. Nuerxiati Abudurexiti contributes to the mathematical modeling, analytical derivations, and theoretical justification of the methodology. Zhengyong Jiang participated in data analysis. Angelos Stefanidis participated in drafting the manuscript. Hongbin Liu and Jionglong Su, as corresponding authors, jointly participated in determining the experimental design and providing technical support. All authors participated in contributing to text and the content of the manuscript, including revisions and edits. All authors approve of the content of the manuscript and agree to be hold accountable for the work.

\section{Availability of data and materials}\label{Availability of data and materials}
The data used in this experiment are financial data, without ethical or moral issues involved. The data supporting the findings of this study are available in Yahoo Finance at https://www.finance.yahoo.com. These data were derived from the following resources available in the public domain: Python package yfinance.



\bibliography{sn-bibliography}

\end{document}